\documentclass[10pt,twocolumn,letterpaper]{article}

\usepackage[pagenumbers]{cvpr} %

\usepackage{graphicx}
\usepackage{amsmath}
\usepackage{amssymb}
\usepackage{booktabs}

\usepackage[pagebackref,breaklinks,colorlinks]{hyperref}

\usepackage[capitalize]{cleveref}
\crefname{section}{Sec.}{Secs.}
\Crefname{section}{Section}{Sections}
\Crefname{table}{Table}{Tables}
\crefname{table}{Tab.}{Tabs.}

\newcommand{\ourtitle}[0]{Benchmarking Self-Supervised Learning on Diverse Pathology Datasets} 
\newcommand{\ourpaperid}[0]{8575}
\newcommand{\ourauthors}[0]{
Mingu Kang\footnotemark[1] \hspace{3mm}
Heon Song\footnotemark[1] \hspace{3mm}
Seonwook Park \hspace{3mm}
Donggeun Yoo \hspace{3mm}
S{\'e}rgio Pereira \\[1mm]
Lunit Inc. \\[0.5mm]
{\tt\small \{jeffkang, heon.song, spark, dgyoo, sergio\}@lunit.io
}
\vspace{-3mm}
}

\usepackage{verbatim}
\usepackage{dsfont}
\usepackage{cancel}
\usepackage{paralist}
\usepackage{subcaption}
\usepackage{stix}
\usepackage{suffix}

\usepackage[usenames,dvipsnames]{xcolor}
\newcommand\authorcomment[3]{\noindent\textsf{\textcolor{#1}{[\textbf{#2:} \textit{#3}]}}}
\WithSuffix\newcommand\authorcomment*[2]{\noindent\textcolor{#1}{\textit{#2}}}
\newcommand{\jeff}[1]{\authorcomment{Orange}{Mingu}{#1}}
\WithSuffix\newcommand\jeff*[1]{\authorcomment*{Orange}{#1}}
\newcommand{\hsong}[1]{\authorcomment{Maroon}{Heon}{#1}}
\WithSuffix\newcommand\hsong*[1]{\authorcomment*{Maroon}{#1}}
\newcommand{\swpark}[1]{\authorcomment{Magenta}{Seonwook}{#1}}
\WithSuffix\newcommand\swpark*[1]{\authorcomment*{Magenta}{#1}}

\newcommand{\blockcomment}[1]{}
\usepackage{comment}

\newcommand\parahead[1]{\vspace{2mm}\noindent\textbf{#1.}\medspace}
\WithSuffix\newcommand\parahead*[1]{\vspace{2mm}\noindent\textbf{#1}\medspace}

\usepackage[normalem]{ulem}
\useunder{\uline}{\ul}{}

\newcommand{\pretrain}[0]{pre-train}
\newcommand{\moco}[0]{MoCo v2\xspace}
\newcommand{\bt}[0]{Barlow Twins\xspace}
\newcommand{\pcam}[0]{PCam\xspace}
\newcommand{\lnt}[0]{TULIP\xspace}

\def\cvprPaperID{\ourpaperid} %
\def\confName{CVPR}
\def\confYear{2023}

\begin{document}

\title{\ourtitle\vspace{-2mm}}
\author{\ourauthors}

\twocolumn[{%
\renewcommand\twocolumn[1][]{#1}%
\maketitle
\begin{center}
    \centering
    \captionsetup{type=figure}  
    \vspace*{-5mm}
    \includegraphics[width=\linewidth]{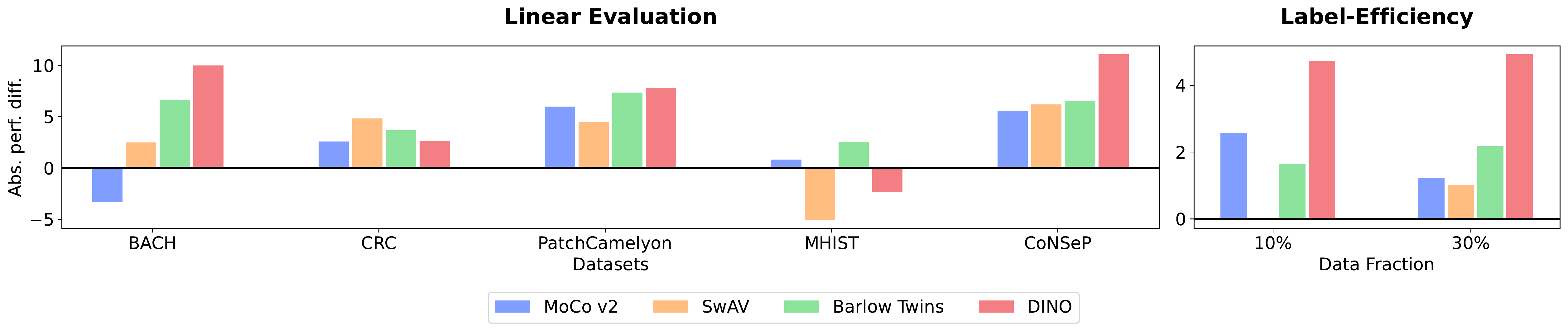}
    \vskip -2mm
    \captionof{figure}{
        \textbf{Self-supervised \pretrain ing on pathology data improves performance on pathology downstream tasks} compared to ImageNet-supervised baselines.
        The $y$-axes show absolute differences in downstream task performance (Top-1 Acc. or mPQ Score).
        Linear evaluation (\textbf{left}) is performed on 4 classification tasks (BACH, CRC, PatchCamelyon, and MHIST) and 1 nuclei instance segmentation task (CoNSeP).
        Label-efficiency (\textbf{right}) is assessed by fine-tuning using small fractions of labeled data from the CoNSeP dataset.
        \label{fig:teaser}
    }
\end{center}
}]

\newcommand{\customfootnotetext}[2]{{
\renewcommand{\thefootnote}{#1}
\footnotetext[0]{#2}}}
\customfootnotetext{*}{The first two authors contributed equally.}

\begin{abstract}
   \vspace*{-3mm}
   Computational pathology can lead to saving human lives, but models are annotation hungry and pathology images are notoriously expensive to annotate.
Self-supervised learning (SSL) has shown to be an effective method for utilizing unlabeled data, and its application to pathology could greatly benefit its downstream tasks.
Yet, there are no principled studies that compare SSL methods and discuss how to adapt them for pathology.
To address this need, we execute the largest-scale study of SSL \pretrain ing on pathology image data, to date.
Our study is conducted using 4 representative SSL methods on diverse downstream tasks.
We establish that large-scale domain-aligned \pretrain ing in pathology consistently out-performs ImageNet pre-training in standard SSL settings such as linear and fine-tuning evaluations, as well as in low-label regimes.
Moreover, we propose a set of domain-specific techniques that we experimentally show leads to a performance boost.
Lastly, for the first time, we apply SSL to the challenging task of nuclei instance segmentation and show large and consistent performance improvements.  %
We release the \pretrain ed model weights\footnote{\tiny\url{https://lunit-io.github.io/research/publications/pathology_ssl}}.
   \vspace*{-5mm}
\end{abstract}

\section{Introduction}
The computational analysis of microscopic images of human tissue -- also known as computational pathology -- has emerged as an important topic of research, as its clinical implementations can result in the saving of human lives by improving cancer diagnosis~\cite{spanhol2015dataset} and treatment~\cite{park2022artificial}.
Deep Learning and Computer Vision methods in pathology allow for objectivity~\cite{choi2022artificial}, large-scale analysis~\cite{diao2021human},
and triaging~\cite{campanella2019clinical} but often require large amounts of annotated data~\cite{van2021deep}.
However, %
the annotation of pathology images requires specialists with many years of clinical residency~\cite{litjens20181399}, 
resulting in scarce labeled public datasets and the need for methods to train effectively on them.

When annotated data is scarce for a given Computer Vision task, one common and practical solution is to fine-tune a model that was \pretrain ed in a supervised manner using the ImageNet dataset~\cite{deng2009imagenet,kornblith2019better}.
This paradigm of transfer learning~\cite{kornblith2019better} was recently challenged by self-supervised learning (SSL), which trains on large amounts of unlabeled data only, yet out-performs supervised \pretrain ing on ImageNet~\cite{chen2020simple,grill2020bootstrap,caron2021emerging}.
In the field of pathology, large unlabeled datasets are abundant~\cite{weinstein2013cancer,litjens20181399,lonsdale2013genotype,bulten2022artificial} in contrast to the lack of annotated datasets~\cite{van2021deep}.
If we were to apply SSL effectively to this huge amount of unlabeled data, downstream pathology tasks could benefit greatly even if they contain limited amount of annotated training data.
Naturally, we ask the question: \emph{How well does self-supervised learning help in improving the performance of pathology tasks?}

ImageNet \pretrain ed weights are commonly used in medical imaging and are known to be helpful in attaining high task performance~\cite{iglovikov2018ternausnet,ke2021chextransfer,xie2018pre, raghu2019transfusion}.
Due to the difference between natural images and medical images, large-scale domain-aligned \pretrain ing has the potential to push performance beyond ImageNet \pretrain ing~\cite{matsoukas2022makes}.
Accordingly, recent works show that SSL \pretrain ing on pathology data can improve performance on downstream pathology tasks~\cite{ciga2022self,gildenblat2019self,boyd2021self,wang2021transpath}.
Our study aims to expand on these previous works by evaluating multiple SSL methods on diverse downstream pathology tasks.
In addition, we propose techniques to adapt SSL methods that were designed for natural image data, to better learn from pathology data.

To understand how to adapt 
existing SSL methods to work on pathology image data, we must identify several key differences between natural and pathology imagery.
Unlike natural images, pathology images can be rotated arbitrarily (impossible to determine a canonical orientation) and exhibit fewer variations in color.
Also, pathology images can be interpreted differently depending on the field-of-view (FoV) due to the multiple hierarchies and contextual differences involved in each task.
We propose to overcome these differences when adapting SSL methods for pathology data, via changes to the training data augmentation scheme in particular, during \pretrain ing.

In this paper, we carry out an in-depth analysis of 4 recent and representative SSL methods; \moco \cite{chen2020improved}, SwAV \cite{caron2020unsupervised}, \bt \cite{zbontar2021barlow}, and DINO \cite{caron2021emerging}, when applied to large-scale pathology data.
For this purpose, we source 19 million image patches from Whole Slide Images (WSI) in The Cancer Genome Atlas (TCGA) dataset~\cite{weinstein2013cancer} and apply our domain-specific techniques in training the SSL methods on this data.
The evaluations are conducted for 2 different downstream tasks over 5 datasets: (1) pathological image classification using BACH~\cite{aresta2019bach}, CRC~\cite{kather_jakob_nikolas_2018_1214456}, MHIST~\cite{wei2021petri}, and PatchCamelyon~\cite{veeling2018rotation} datasets, and (2) nuclei instance segmentation and classification using the CoNSeP dataset \cite{graham2019hover}.

Our large-scale study yields several useful contributions:
\begin{inparaenum}[(a)]
    \item we conduct the largest-scale study of SSL \pretrain ing on pathology image data to date, and show its benefit over using ImageNet \pretrain ed weights on diverse downstream tasks (see~\autoref{fig:teaser}), 
    \item we propose a set of carefully designed data curation and data augmentation techniques that can further improve downstream performance,
    \item we demonstrate that SSL is label-efficient, and is therefore a practical solution in pathology where gathering annotation is particularly expensive, and
    \item for the first time, we apply SSL to the dense prediction task of nuclei instance segmentation and show its value under diverse evaluation settings.  %
\end{inparaenum}
We release our \pretrain ed model weights at {\small \url{https://lunit-io.github.io/research/publications/pathology_ssl}}
to further contribute to the research community.

\section{Related Work}

\subsection{Self-supervised Representation Learning}
SSL methods learn representations through pre-text tasks designed to exploit supervisory signals obtained from the unlabeled data itself.
We describe the 4 major paradigms of SSL as commonly discussed in literature.

\parahead{Contrastive Learning} 
Contrastive methods~\cite{he2020momentum,misra2020self,oord2018representation} such as SimCLR \cite{chen2020simple} and \moco \cite{chen2020improved} learn to discriminate each training data instance from all the others. 
The objective is to learn similar representations of positive pairs (perturbations by data augmentation) and discriminative representations in relation to negative pairs (other instances). 
A limitation is the need for diverse negative pairs, which is mitigated through large batch sizes \cite{chen2020simple} or memory banks \cite{chen2020improved}. 
In this work, we explore \moco \cite{chen2020improved}.
 
\parahead{Non-contrastive Learning} 
Methods such as BYOL \cite{grill2020bootstrap}, SimSiam \cite{chen2021exploring}, and \bt \cite{zbontar2021barlow}, share similarities with \textit{contrastive learning} methods in that they learn representations of images under different augmented views.
The fundamental difference is that these approaches do not rely on negative pairs, which allows them to work with small batch sizes. 
In this work, we explore \bt \cite{zbontar2021barlow}.

\parahead{Clustering} 
This paradigm uses the concept of clustering and is shown in DeepCluster~\cite{caron2018deep} and SwAV~\cite{caron2020unsupervised}.
Clustering-based SSL discriminates between clusters of image representations instead of explicit pairs of images. 
In this work, we explore SwAV \cite{caron2020unsupervised}.

\parahead{SSL with VisionTransformer}
The effectiveness of Vision Transformers (ViT)~\cite{dosovitskiy2020image} has been demonstrated on various computer vision tasks.
Thus, the paradigm shift from CNN to ViT has recently emerged in the field of self-supervised learning.
Consequently, recent studies~\cite{chen2021empirical, caron2021emerging, li2021efficient} try to investigate techniques that facilitate SSL with ViT-based architectures.
In this work, we explore DINO \cite{caron2021emerging}.

\graphicspath{ {./sections/images/} }

\begin{figure*}[th]
    \centering
    \vspace*{-2.5mm}
    \includegraphics[width=0.93\textwidth]{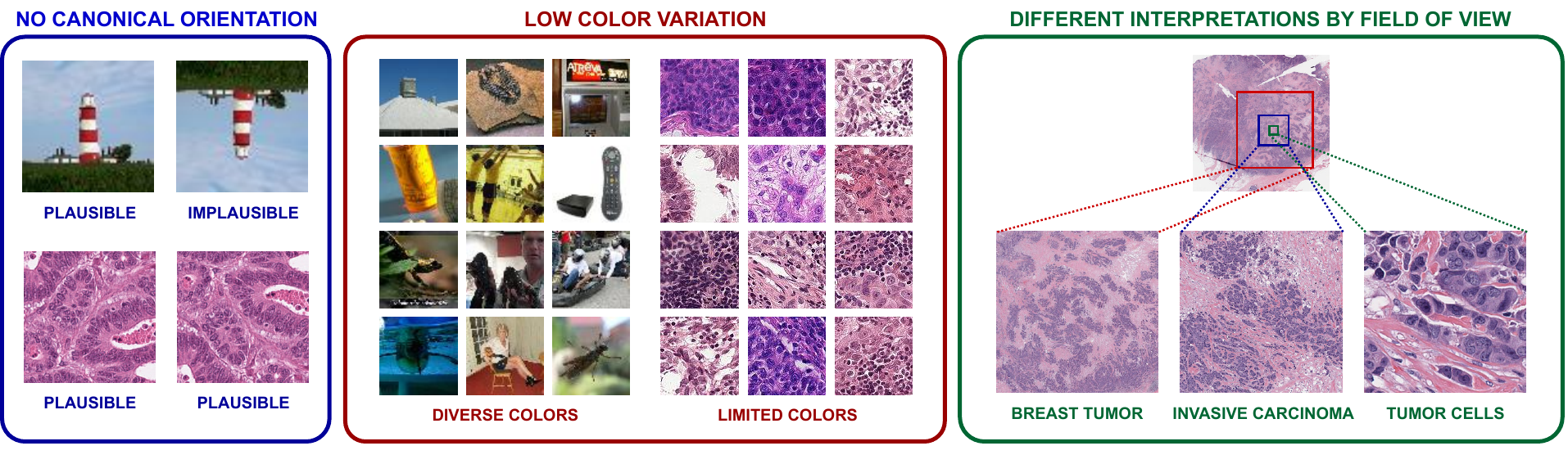}
    \vspace*{-2.5mm}
    \caption{
    \textbf{Pathology images vs natural images.}
    Pathology images are different from natural images in 3 major ways. They have no canonical orientation (no ``right way up''), have low color variation, and can be interpreted differently depending on field-of-view.
    Consequently, self-supervised learning methods need to be implemented differently when working in the domain of pathology.
    }
    \vspace*{-3.0mm}
    \label{fig:image_cmp}
\end{figure*}

\subsection{SSL in Medical Imaging}

Recently, \cite{matsoukas2022makes} investigates transfer learning in medical imaging and observes that using domain-aligned datasets for \pretrain ing improves the transferability of models.
Moreover, domain-specific SSL methods can further improve the performance of models fine-tuned on downstream medical image-related tasks ~\cite{sowrirajan2021moco, azizi2021big, boyd2021self, wang2021transpath, ciga2022self, gildenblat2019self, yang2022concl}.
In pathology, \cite{wang2021transpath} employs BYOL and evaluates \pretrain ed weights learned from pathology data on image classification tasks.
\cite{gildenblat2019self} adopts SimSiam, showing that SSL improves pathology image retrieval.
Also recently, \cite{ciga2022self} uses SimCLR and observes that SSL consistently improves on downstream pathology tasks compared to ImageNet pre-training. %

Unlike previous works that focus on just a single SSL approach~\cite{ciga2022self,chen2022scaling,li2021dual}, or either CNNs or ViTs only~\cite{yang2022concl}, we explore one representative method from each of the aforementioned SSL paradigms including ViT-based SSL. In this way, we establish a common and fair benchmark for comparing these methods in pathology.
Furthermore, we evaluate the domain-specific \pretrain ed weights on various downstream tasks, including the challenging task of nuclei instance segmentation.
Finally, we devise techniques for data augmentation that are specifically tailored to tackle pathology-specific challenges, thus leading to better representations and performance in the downstream tasks.

\section{Self-supervised Pre-training for Pathology}

The performance of SSL methods can vary greatly depending on the composition of training data and the selected set of data augmentation methods.
SSL methods in the literature are commonly designed and evaluated in settings involving natural images and may benefit from further adaptation when applied to different domains, such as pathology.
In this section, we discuss the differences between natural images and pathology images. We also propose a set of techniques that can be easily adopted to improve the performance of models \pretrain ed on pathology image data.

\subsection{Differences to Natural Images}
\label{method:Adaptiong_SSL}

Images contained in popular Computer Vision datasets (e.g. ImageNet~\cite{deng2009imagenet}) are often denoted as ``natural images''.
Pathology images have several unique characteristics that make them distinct from natural images.
We discuss these differences in this section and summarize them in~\autoref{fig:image_cmp}.

\parahead{No canonical orientation}
Objects or scenes contained in natural images are oriented based on plausibility, i.e. how a human expects the objects to be oriented.
Methods in Computer Vision can take advantage of such assumptions or patterns (e.g. Manhattan World assumption~\cite{coughlan2000manhattan}) and thus SSL methods do not randomly augment the orientation of images at training time.
However, pathology images can be oriented in any way and still remain plausible.
Furthermore, objects (e.g. cells) are many and dispersed at arbitrary locations, 
making it impossible to define a ``canonical orientation'', i.e. the correct standard orientation.

\parahead{Low color variation}
While natural images contain a large range of colors due to the diversity of represented objects, pathology images tend to display similar color distributions (e.g. purple and pink staining).
Though the  staining can vary across institutions and the same biological structures have different appearances depending on the cancer type, pathology images are more consistent than natural images.

\parahead{Different FoVs}
To correctly analyze pathology images, different Field of Views (FoVs) must be considered.
A larger FoV allows pathologists and algorithms to better understand the larger context of the tissue regions and cell classes to make high-level predictions such as the grading of prostate cancer~\cite{bulten2022artificial}.
In other tasks that require the classification of individual cells or communities of cells, a smaller FoV is required to increase the resolution on the objects of interest~\cite{graham2019hover}.
Thus, a \pretrain ed model for pathology should ideally be able to handle tasks from diverse FoVs.

\subsection{Techniques to Adapt SSL for Pathology}
\label{method:technique_for_SSL}
In this section, we introduce our techniques for adapting SSL methods for pathology imagery.

\parahead{Random vertical flips}
Unlike natural images, pathology images are no less plausible or realistic when they are vertically flipped. We therefore propose to randomly apply vertical flips during SSL \pretrain ing.

\parahead{Stain augmentations}
The typical color distortion employed by SSL methods applies a strong amount of jitter to brightness, contrast, and saturation, resulting in pathology images that look highly unrealistic.
\cite{tellez2019quantifying} proposes to apply this jitter in pathology-specific color spaces such as the HED-space~\cite{ruifrok2001quantification}.
\cite{shen2022randstainna} points out that naive jittering can produce unrealistic images and proposes RandStainNA.
RandStainNA fits unimodal Gaussian distributions to the channel-wise statistics of 3 color spaces (HSV, Lab, HED) using images from the training dataset.
At training time, a color space is randomly chosen, then the target channel-wise mean and standard deviations for that color space are sampled from the fitted Gaussian distributions.
Reinhard's method~\cite{reinhard2001color} is used to re-normalize the input image to match the target statistics.
RandStainNA is shown to improve supervised learning performance for pathology, and therefore we adopt it for our SSL \pretrain ing.

\begin{figure}
    \centering
    \begin{tabular}{@{}m{0.07\columnwidth}@{}m{0.91\columnwidth}@{}}
        (a) & \includegraphics[width=0.91\columnwidth]{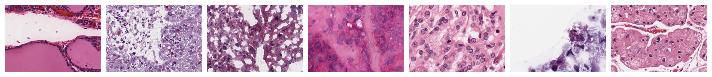} \\[-1.2mm]
        (b) & \includegraphics[width=0.91\columnwidth]{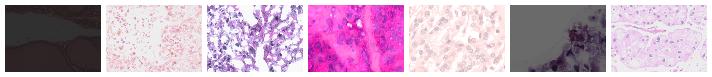} \\[-1.2mm]
        (c) & \includegraphics[width=0.91\columnwidth]{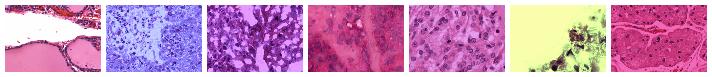} \\[-1.2mm]
        (d) & \includegraphics[width=0.91\columnwidth]{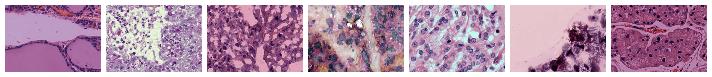} \\[-1.2mm]
        (e) & \includegraphics[width=0.91\columnwidth]{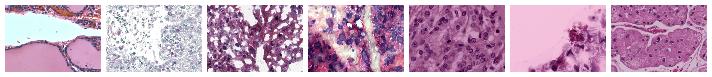} \\[-1.2mm]
    \end{tabular}
    \vspace*{-2.5mm}
    \caption{
    \textbf{Color augmentations on pathology images.}
    (a) input images, (b) color jitter as used in typical SSL methods~\cite{chen2020simple}, (c) HED-light~\cite{tellez2019quantifying}, (d) RandStainNA~\cite{shen2022randstainna}, and (e) RandStainNA$_{GMM}$.
    RandStainNA and RandStainNA$_{GMM}$ can produce more realistic and plausible pathology images.
    }
    \vspace*{0mm}
    \label{fig:aug_images}
\end{figure}

Furthermore, we attempt to improve the realism of RandStainNA by fitting a Gaussian Mixture Model (GMM) with 10 components, to the channel-wise statistics of each color space.
The GMM can fit the covariances between variables and respect the multi-modality of the channel-wise mean and standard deviation values.
We denote this as RandStainNA$_{GMM}$ and show the visual differences against alternative methods in \autoref{fig:aug_images}.

Lastly, we observe that previous works in SSL \cite{chen2020simple, zbontar2021barlow, grill2020bootstrap, chen2021exploring} highlight the importance of color distortion.
We therefore propose to apply color distortion with a weaker jittering strength as done in~\cite{ciga2022self}.
Our main experiments adopt RandStainNA$_{GMM}$ along with random grayscale and a weaker color jittering augmentation.

\parahead{Using multiple FoVs}
As aforementioned, some pathology tasks require high FoV while others benefit from low FoV.
We identify that pathology tasks (e.g., image classification and instance segmentation)
are commonly defined at approximately $20\times$~\cite{aresta2019bach,kather_jakob_nikolas_2018_1214456} or $40\times$~\cite{graham2019hover,veeling2018rotation,wei2021petri} objective magnification.
Therefore, we build our large-scale unlabeled dataset using image patches from both magnifications.

\section{Experiment Setup}

\begin{table}[t]
\vspace*{0mm}
\small\centering
\renewcommand{\arraystretch}{0.9}
\begin{tabular}{llccc}
\toprule
\multirow{2}{9mm}{\textbf{Data source}} & \multirow{2}{8.5mm}{\textbf{No. of WSIs}} &  \multicolumn{3}{c}{\textbf{No. of patches}} \\ \cmidrule(lr){3-5}
& & \multicolumn{1}{c}{20x} & 40x & Total \\ \midrule
TCGA &  20,994 & \multicolumn{1}{c }{9,497,768} & \multicolumn{1}{c}{9,502,301} & 19,000,069 \\ 
\lnt & 15,672 & \multicolumn{1}{c}{7,084,130} & \multicolumn{1}{c}{6,494,358} & 13,578,488 \\ \midrule
Total & 36,666 & 16,581,898 & 15,996,659 & 32,578,557 \\ \bottomrule
\end{tabular}\vskip -2mm
\caption{\textbf{Unlabeled data for pre-training}. Amount of data used for \pretrain ing in terms of the number of WSIs and the actual number of extracted patches (per FoV).
}
\vspace*{-3mm}
\label{method:pretraining_dataset}
\end{table}
\subsection{Pre-training Dataset}
\autoref{method:pretraining_dataset} presents the scale of unlabeled data used for \pretrain ing.
We first collect 20,994 WSIs from The Cancer Genome Atlas (TCGA) and 15,672 WSIs from \lnt. Both datasets consist of Hematoxylin \& Eosin (H\&E) stained WSIs of various cancers. TCGA is publicly available and widely used for training deep learning models~\cite{cruz2017accurate,diao2021human,park2022artificial}.
\lnt~is an internally collected dataset. %
To increase diversity and keep our experimental setting practical, we extract at most 1,000 patches of resolution 512 $\times$ 512 pixels from each slide, resulting in a total of 32.6M patches (19M from TCGA and 13.6M from TULIP).
The \pretrain ing data covers two different FoVs; 20$\times$ ($0.5 \mu$m/px) and 40$\times$ ($0.25 \mu$m/px) objective magnification.
All experiments, unless specified otherwise, present the results of \pretrain ing on the TCGA dataset only.

\subsection{Downstream Datasets}

\begin{table}[t]
\setlength{\tabcolsep}{3.5pt}
\small\centering
\begin{tabular}{lccccc}
\toprule
\textbf{Dataset} & \textbf{\# Classes} & \textbf{\# Patches} & \textbf{Patch size}       & \textbf{FoV} & \textbf{Task}       \\ \midrule
BACH     & 4         & 400       & 2048$\times$1536 & 20$\times$ & Cls\\
CRC      & 9         & 107,180   & 224$\times$224   & 20$\times$ & Cls\\
PCam     & 2         & 327,680   & 96$\times$96     & 40$\times$ & Cls\\
MHIST    & 2         & 3,152     & 224$\times$224   & 40$\times$ & Cls\\
CoNSeP   & 7         & 41        & 1000$\times$1000 & 40$\times$ & Seg \\ \bottomrule
\end{tabular}\vskip -2mm
\caption{\textbf{Datasets for downstream tasks}.
Note that, Cls indicates ``image classification" and Seg is ``nuclei instance segmentation''.
}
\vspace*{-3mm}
\label{table:downstream_dataset}
\end{table}

We validate the \pretrain ed models under classification and segmentation tasks using various downstream datasets described in~\autoref{table:downstream_dataset}. For image classification, the following four datasets are used: BACH (four-class breast cancer type classification)~\cite{aresta2019bach}, CRC (nine-class human colorectal cancer type classification)~\cite{kather_jakob_nikolas_2018_1214456}, MHIST (binary class colorectal polyp type classification)~\cite{wei2021petri}, and \pcam~(binary class breast cancer type classification)~\cite{veeling2018rotation}. The patches of the datasets are labeled according to the predominant cancer type or the presence of cancers. 
For nuclei instance segmentation, we use CoNSeP~\cite{graham2019hover} which contains segmentation masks for each cell nucleus along with nuclei types.
Further details of the downstream datasets are shown in the supplementary materials.

\begin{table*}[t]
\centering
\setlength{\tabcolsep}{3pt}
\renewcommand{\arraystretch}{0.9}
\begin{tabular}{@{}lllccccccc@{}}
\toprule
\multirow{2}{*}{\textbf{Arch.}} & \multirow{2}{*}{\textbf{Method}} & \multicolumn{2}{c}{\textbf{BACH}} & \multicolumn{2}{c}{\textbf{CRC}} & \multicolumn{2}{c}{\textbf{\pcam}} & \multicolumn{2}{c}{\textbf{MHIST}} \\
&                                  & Linear          & Fine-tune       & Linear          & Fine-tune      & Linear           & Fine-tune       & Linear           & Fine-tune       \\ \midrule
\multirow{5}{*}{ResNet-50}
& \textit{Random}                  & 51.67           & 61.67           & 68.91           & 89.99          & 76.52            & 75.71           & 63.15            & 75.54           \\
& \textit{Supervised}              & 80.83           & {\ul 86.67}     & 90.93           & 92.09          & 80.79            & 80.63           & 76.25            & 78.92           \\
& \moco                            & 77.50           & \textbf{90.83}  & 93.52           & \textbf{96.21} & {\ul 86.78}      & \textbf{87.62}  & {\ul 77.07}      & \textbf{85.88}  \\
& SwAV                             & {\ul 83.33}     & 82.50           & \textbf{95.78}  & {\ul 93.31}    & 85.28            & {\ul 87.60}     & 71.14            & 77.99           \\
& BT                               & \textbf{87.50}  & 85.00           & {\ul 94.60}     & 93.23          & \textbf{88.15}   & 86.92           & \textbf{78.81}   & {\ul 81.27}     \\
\midrule
\multirow{4}{*}{ViT-S}
& \textit{Random$_{p=16}$}         & 45.00           & 57.50           & 69.90           & 86.10          & 74.43            & 75.42           & 63.46            & 62.13           \\
& \textit{Supervised$_{p=16}$}     & 75.83           & 85.83           & 91.56           & {\ul 95.81}    & 80.96            & 88.30           & \textbf{78.51}   & \textbf{81.68}  \\
& DINO$_{p=16}$                    & \textbf{85.83}  & {\ul 87.50}     & {\ul 94.19}     & {\ul 95.81}    & {\ul 88.78}      & {\ul 90.40}     & 76.15            & {\ul 79.43}     \\
& DINO$_{p=8}$                     & {\ul 83.33}     & \textbf{93.33}  & \textbf{95.29}  & \textbf{97.13} & \textbf{90.12}   & \textbf{90.76}  & {\ul 77.89}      & 78.40           \\ 
\bottomrule
\end{tabular}
\vskip -2mm
\caption{\textbf{Downstream evaluation of the image classification tasks}. We report Top-1 accuracy for both linear and fine-tuning experiment protocols for models trained using the TCGA data source. Note that, $p$ represents the patch size used in ViT.}
\vspace*{-3mm}
\label{result:qualitative_classification_results}
\end{table*}

\subsection{Pre-training Details}
We learn representations using 4 different SSL methods. 
Unless otherwise mentioned, we use the ResNet-50 ($1\times$) \cite{he2016deep} architecture for \moco \cite{chen2020improved}, \bt \cite{zbontar2021barlow}, and SwAV \cite{caron2020unsupervised}.
For DINO \cite{caron2021emerging}, we use ViT-Small \cite{dosovitskiy2020image} with different patch sizes, 16$\times$16 and 8$\times$8 (denoted DINO$_{p=16}$ and DINO$_{p=8}$, respectively), as it has a comparable number of parameters to ResNet-50 ($1\times$). %
We follow the proposed recipe of each SSL method and launch the \pretrain ing, distributed across 64 NVIDIA V100 GPUs.
The linear scaling rule \cite{goyal2017accurate} is applied to adjust the learning rate: \textit{lr} = $lr_{method} * batchsize/256$.
We adopt the concept of the \emph{``ImageNet epoch''} from \cite{tian2021divide} for ease of analysis and train models for 200 ImageNet epochs, across all experiments. 
We define 10 ImageNet epochs for the warmup, and the cosine scheduler is followed.
The details of configurations can be found in supplementary materials. %

\subsection{Downstream Training Details}
For the downstream tasks, we follow the standard practice as introduced in various SSL papers~\cite{chen2020simple,grill2020bootstrap,zbontar2021barlow}. 
For image classification tasks, the datasets are split into training, validation, and test sets. 
We perform the hyper-parameter search based on the validation set, reporting the performance on the test set.
For the segmentation task, the Hover-Net \cite{graham2019hover} architecture -- a standard architecture in the nuclei instance segmentation task -- is adopted with the \pretrain ed backbone. We follow the same data pre-processing and training schemes as in \cite{graham2019hover} to enable reproducibility and fair comparison of results. Further details of downstream tasks can be found in the supplementary materials.

\subsection{Evaluation Metrics}
For image classification, we report top-1 accuracy, while using panoptic quality (PQ)  \cite{kirillov2019panoptic} for nuclei instance segmentation.
PQ is a standard metric for assessing the performance of nuclear instance segmentation \cite{graham2019hover} that accounts for both detection and segmentation quality with respect to each instance. The PQ metric is defined as,
\begin{equation}
    PQ = \frac{\sum_{(p,g)\in TP} \mathrm{IoU}{(p,g)}}{|TP| + \frac{1}{2}|FP| + \frac{1}{2}|FN|},
\end{equation}
where $p$ denotes a predicted mask for each nuclei class and $g$ denotes a corresponding ground truth. 
The numerator $\sum_{(p,g)\in TP} \mathrm{IoU}{(p,g)}$ represents the summation of correctly matched Intersection Over Union (IoU) over all pairs between predictions and ground truth. 
We count pairs between predictions and ground truths with IoU of more than 0.5 as True Positives (TP).
False Positives (FP) and False Negatives (FP) represent wrongly predicted predictions and ground truth, respectively.
Note that we use multi-class PQ (mPQ) to measure the performance of instance segmentation and classification simultaneously.

\section{Experimental Results}

In this section, we carry out various experiments on the downstream tasks of image classification and nuclei instance segmentation.
Through these experiments, we compare the utility of various SSL pre-training methods in the light of downstream pathology task performance.
First, we present our quantitative results using well-established evaluation protocols.
We then demonstrate the benefit of SSL \pretrain ing with a limited number of labeled data and under different fine-tuning schedules.
Finally, we perform an ablation study to quantitatively verify the efficacy of our techniques for adapting SSL methods to pathology data.

In evaluating downstream task performance, we stick to well-established evaluation protocols in SSL.
The first is \emph{linear evaluation} (denoted \emph{Linear}), where we freeze the backbone and train the remaining parts of the model (e.g., linear classifier or decoders).
The second is \emph{full fine-tuning} (denoted \emph{Fine-tune}), where all layers including the backbone are fine-tuned.
The former protocol assesses the quality of learned representations, whereas the latter evaluates the transferability of learned weights. %
In our experiments, we compare against the ImageNet-supervised (denoted \textit{Supervised}) \pretrain ing baseline of the corresponding backbone type as well as a random initialization (denoted \textit{Random}) baseline.

\subsection{Image Classification}

We present our linear evaluation and fine-tuning results for 4 image classification benchmarks in \autoref{result:qualitative_classification_results}.

\parahead{Linear evaluation}
In linear evaluation results, we find that self-supervised TCGA pre-training typically out-performs supervised ImageNet pre-training.
Of the ResNet-50 based SSL methods, Barlow Twins performs consistently well, out-performing other methods on the BACH, \pcam, and MHIST datasets.
Of the ViT-based SSL methods, DINO$_{p=16}$ achieves competitive results, and DINO$_{p=8}$ performs even better on the CRC and PCam datasets.
The improved performance of DINO$_{p=8}$ is in line with observations from \cite{caron2021emerging} which shows that performance improves at the cost of computation with smaller patch size.
One exception is on the MHIST dataset, where the supervised baseline shows good performance.
Based on the linear evaluation results, we can certainly claim that domain-aligned pre-training improves representation quality.

\parahead{Fine-tuning}
Under fine-tuning, we find that the trends are similar but with notable changes.
Firstly, as shown in other SSL works, the performance gap between ImageNet-supervised and TCGA-SSL reduces.
Furthermore, MoCo~v2 shows consistently high performance among CNN methods, showing that it may be the architecture of choice for transfer learning settings using CNNs.
Regarding ViTs,
we find that the trends are almost identical to linear evaluation except that the fine-tuned performances are often better than CNN counterparts trained using SSL on TCGA data.
For classification tasks, then, SSL using ViT on large-scale pathology data is beneficial.

\begin{table}[t]
\centering
\setlength{\tabcolsep}{3pt}
\renewcommand{\arraystretch}{0.80}
\begin{tabular}{@{}llcc@{}}
\toprule
\multirow{2}{*}{\textbf{Arch.}} & \multirow{2}{*}{\textbf{Method}} & \multicolumn{2}{c}{\textbf{CoNSeP}}   \\
                                &                                  & Linear            & Fine-tune         \\ \midrule
\multirow{5}{*}{ResNet-50}      & \textit{Random}                  & 22.29             & 46.72             \\
                                & \textit{Supervised}              & 34.25             & 49.60             \\
                                & \moco                            & 39.85             & \textbf{51.75}    \\
                                & SwAV                             & \underline{40.45} & 51.16             \\
                                & BT                               & \textbf{40.79}    & \underline{51.61} \\ \midrule
\multirow{4}{*}{ViT-S}          & \textit{Random$_{p=16}$}         & 20.55             & 27.19             \\
                                & \textit{Supervised$_{p=16}$}     & 21.43             & 36.70              \\
                                & DINO$_{p=16}$                    & \underline{32.54} & \underline{38.43}  \\
                                & DINO$_{p=8}$                     & \textbf{42.71}    & \textbf{46.70}     \\ \bottomrule
\end{tabular}
\vskip -2mm
\caption{\textbf{Downstream evaluation for the nuclei instance segmentation task}. We report the mPQ score for both linear and fine-tuning experiment protocols for models trained using the TCGA data source. }
\vspace*{-4mm}
\label{result:qualitative_segmentation _results}
\end{table}

\subsection{Nuclei Instance Segmentation}

To the best of our knowledge, we show for the first time, the effect of self-supervised domain-aligned pre-training on a downstream dense prediction task.
We run experiments on the CoNSeP dataset for the task of nuclei instance segmentation and report the mPQ score in~\autoref{result:qualitative_segmentation _results}.

\parahead{CNN experiments}
The performance of SSL using ResNet-50 shows similar trends to the case of image classification, where Barlow Twins performs well on the linear evaluation protocol and MoCo v2 performs well in fine-tuning.
More consistently than in the case of classification, SSL \pretrain ed models out-perform supervised ImageNet \pretrain ing by a large margin, especially considering the difficulty of increasing the mPQ score.

\parahead{ViT experiments}
To the best of our knowledge, we are the first to integrate ViT backbones into the HoVer-Net architecture for nuclei instance segmentation.
We find that DINO trained on TCGA data out-performs ImageNet-trained weights for DINO$_{p=16}$ by a large margin, showing again the importance of domain-aligned SSL.
While DINO$_{p=16}$ does not work well in neither linear nor fine-tuning evaluations, DINO$_{p=8}$ out-performs even CNN-based methods in linear evaluation and performs reasonably well with fine-tuning.
Future work may be able to further unlock the power of transformers as a backbone for nuclei instance segmentation.

\begin{table}[t]
\centering
\setlength{\tabcolsep}{1.8pt}
\renewcommand{\arraystretch}{0.9}
\resizebox{\linewidth}{!}
{
\begin{tabular}{@{}llccccc@{}}
\toprule
\multirow{2}{*}{\textbf{Method}} & \multirow{2}{8mm}{\textbf{Pre. Data}} & \multicolumn{4}{c}{\textbf{Top-1 Acc.} (\%)} & \textbf{mPQ} \\ \cmidrule(lr){3-6} \cmidrule(lr){7-7}
 &  & BACH & CRC & PCam & MHIST & CoNSeP \\ \midrule
\textit{Random}     & -     &         51.67  &         68.91  &         76.52  &         63.15  &         22.29  \\
\textit{Supervised} & IN    &    {\ul 80.83} &         90.93  &         80.79  &         76.25  &         33.49  \\
\moco               & IN    &         71.67  &         92.86  &         82.37  &    {\ul 79.73} &         39.13  \\
\moco               & TCGA  &         77.50  &    {\ul 93.52} & \textbf{86.78} &         77.07  &    {\ul 39.85} \\
\moco               & TC+TU & \textbf{85.00} & \textbf{93.94} &    {\ul 86.53} & \textbf{82.29} & \textbf{41.40} \\
\bottomrule
\end{tabular}
}
\vskip -2mm
\caption{
\textbf{Varying pre-training datasets under the linear evaluation protocol.} %
We consider ImageNet (IN), TCGA, and TCGA and \lnt combined (TC+TU) as \pretrain ing datasets.
Training with TC+TU results in consistent performance improvements.}
\vspace{-3mm}
\label{result:linear_internal_data}
\end{table}

\subsection{Pre-training on Different Datasets}
The experiment aims to investigate the impact on the downstream task in accordance with the \pretrain ing data.
We select \moco as it has previously shown robust performance in relation to various domain-specific data \cite{van2021revisiting}.
We show our linear evaluation results in~\autoref{result:linear_internal_data} where we compare against \moco pre-training on ImageNet\footnote{We use a publicly available ImageNet \pretrain ed model for \moco.} as well as on the combined TCGA and \lnt data.
We note that TCGA-pretraining out-performs supervised/SSL pre-training with ImageNet on BACH, CRC, PCAM, and CoNSeP.
When adding \lnt data into the mix, we can use a total of 36K slides and 32.6M patches for pre-training, and we see that this results in the overall best performance.
Through these experiments, we conclude that using a domain-aligned dataset such as TCGA is useful, and increasing the scale and diversity of data can further boost performance.

\begin{table}[t]
\centering
\setlength{\tabcolsep}{2.5pt}
\renewcommand{\arraystretch}{0.9}
\begin{tabular}{lcccccc}
\toprule
\multirow{3}{*}{\textbf{Method}} & \multicolumn{3}{c}{\textbf{CRC (Top-1 Acc.)}} & \multicolumn{3}{c}{\textbf{CoNSeP (mPQ)}}                         \\ \cmidrule(lr){2-4} \cmidrule(lr){5-7}
                             & 1\%               & 10\%              & 100\%             & 10\%              & 30\%              & 100\%            \\ \midrule
{\footnotesize ResNet-50}    &                   &                   &                   &                   &                   &                  \\
\textit{Supervised}          & 90.28             & {\ul 93.87}       & 92.09             & 40.01             & 41.92             & 49.60            \\
\moco                        & \textbf{91.73}    & \textbf{95.10}    & \textbf{96.21}    & \textbf{42.59}    & {\ul 43.15}       & \textbf{51.75}   \\
SwAV                         & 89.26             & 92.84             & {\ul 93.31}       & 39.97             & 42.94             & 51.16            \\
BT                           & {\ul 91.23}       & 92.84             & 93.23             & {\ul 41.66}       & \textbf{44.10}    & {\ul 51.61} \\ \midrule
{\footnotesize ViT-S}        &                   &                   &                   &                   &                   &                  \\
\textit{Supervised$_{p=16}$} & 93.15             & 94.76             & \underline{95.81} & 18.49             & 20.92             & 36.70            \\
DINO$_{p=16}$                & \underline{94.03} & \underline{94.92} & \underline{95.81} & \underline{23.22} & \underline{25.85} & \underline{38.43} \\ 
DINO$_{p=8}$                 & \textbf{95.03}    & \textbf{96.27}    & \textbf{97.13}    & \textbf{35.53}    & \textbf{37.82}    & \textbf{46.70}    \\ \bottomrule
\end{tabular}
\vskip -2mm
\caption{\textbf{Label-efficiency.} Full fine-tuning results when using a limited number of training samples for the CRC and CoNSeP downstream benchmarks.
}
\label{result:annotation_study}
\end{table}

\subsection{Fine-tuning with Limited Labeled Data}
In the pathology domain, acquiring high-quality annotations require expert-derived labor and exhaustive refinement to maximize consensus across annotators, hindering the establishment of large-scale annotated datasets.
Prior findings in computer vision and medical imaging show that SSL \pretrain ed models are label-efficient under fine-tuning evaluations \cite{chen2020big, azizi2021big}.
To evaluate the label-efficiency of \pretrain ed models in the pathology domain, we perform similar evaluations and fine-tune our models while varying the labeled data fraction.
Following prior works~\cite{chen2020simple,zbontar2021barlow}, subsets of size 1\% and 10\% are sampled for the image classification dataset.
We pick CRC dataset since it has sufficient amounts of data as well as a number of classes to conduct the experiment.
On the other hand, the CoNSeP dataset has insufficient data to conduct the same experiment, and therefore, we use subsets of size 10\% and 30\% for nuclei instance segmentation. %
Further details can be found in our supplementary materials.

\autoref{result:annotation_study} presents fine-tuning evaluation results with varying amounts of downstream labeled data.
Compared to the \textit{Supervised} baselines, SSL \pretrain ed weights out-perform the ImageNet \pretrain ed weights.
In particular, \moco and DINO$_{p=8}$ show the best performances for ResNet-50 and ViT-S backbones respectively, maintaining the performance gap to \textit{Supervised} baselines even with increasing amounts of labeled data.

\begin{figure}[t]
    \centering
    \vskip -1mm
    \includegraphics[width=1.0\linewidth]{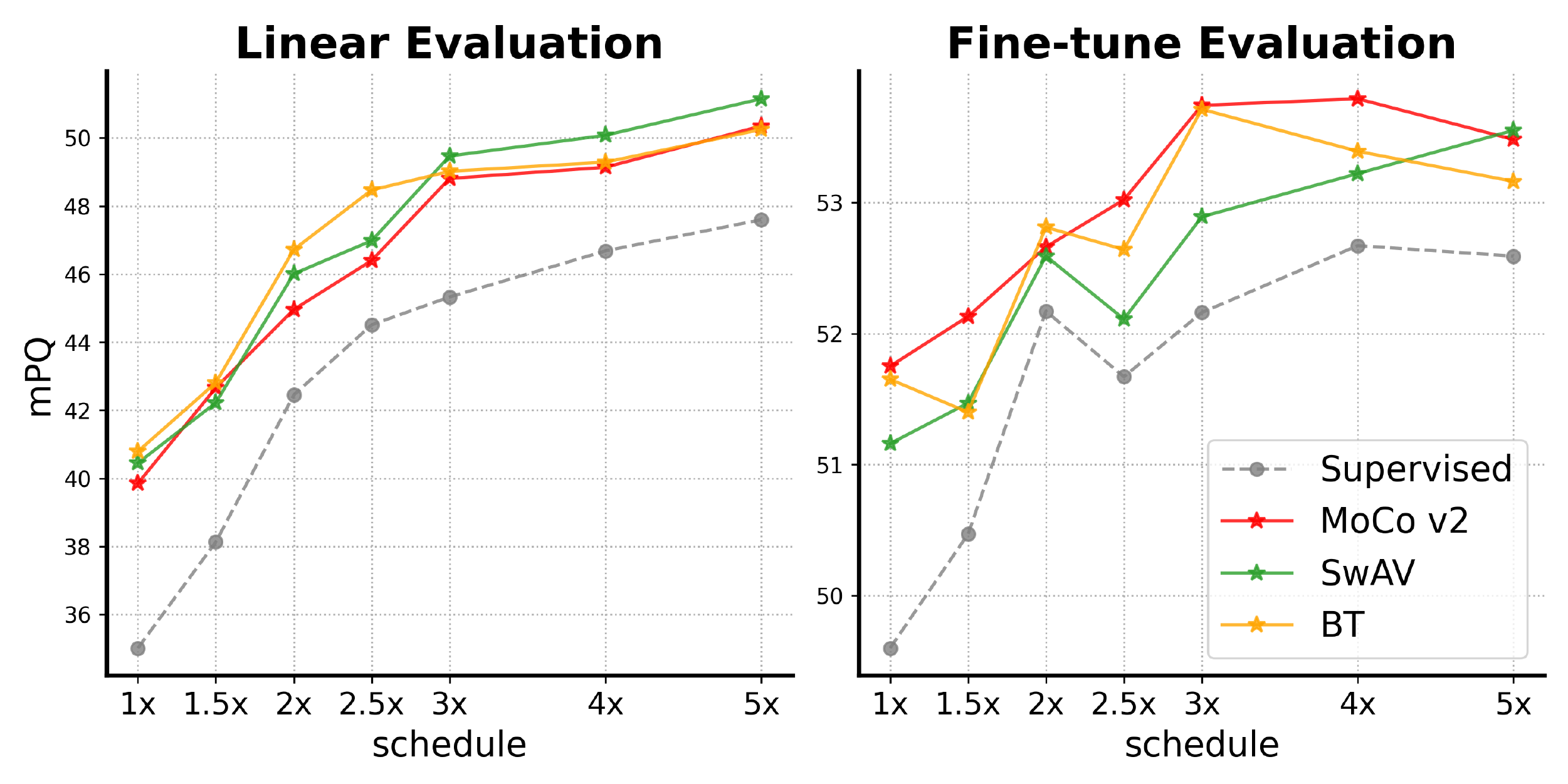}
    \vskip -3mm
    \caption{\textbf{Different learning schedules for the nuclei instance segmentation task using CoNSeP}. We scale up the learning schedule from 1$\times$ (20 epochs) to 5$\times$ (100 epochs).
    }
    \label{result:scheduler_graph}
    \vskip -4mm
\end{figure}

\subsection{Longer Learning Schedules}

When evaluating SSL methods on downstream dense prediction tasks, it is desirable to show results with longer fine-tuning schedules~\cite{he2020momentum,xiao2021region,roh2021spatially} but this is rarely shown in papers due to the diminishing performance gap between methods when fine-tuning for longer~\cite{he2019rethinking}.
To better demonstrate the real-world implications of SSL \pretrain ing on pathology data for nuclei instance segmentation, we scale the training schedule from 1$\times$ to 5$\times$ and show mPQ scores computed on the CoNSeP dataset in~\autoref{result:scheduler_graph}.
In both linear and fine-tuning evaluations, we clearly observe the benefit of TCGA \pretrain ed weights compared to ImageNet-supervised weights and note that the performance gap is maintained even at longer training schedules.

\subsection{Ablation Study}
As described in Sec \ref{method:technique_for_SSL}, we propose to use data augmentation techniques tailored for pathology data.
We empirically assess whether these domain-adapted techniques are beneficial to SSL in pathology through an ablation study.
We select \bt \cite{zbontar2021barlow} to conduct our experiments and \pretrain~our models for 200 ImageNet epochs on the TCGA dataset.
The fine-tuning evaluation protocol is adopted to better indicate real-world implications.

\begin{table}[]
\centering
\renewcommand{\arraystretch}{0.8}
\newcommand{\chk}[0]{\checkmark}
\resizebox{\linewidth}{!}
{
\begin{tabular}{@{}l@{}cccccc@{}}
\toprule
 & \multirow{2}{*}{\textbf{S}} 
 & \multirow{2}{*}{\textbf{V}} 
 & \multirow{2}{*}{\textbf{G}} & \multicolumn{2}{c}{\textbf{ColorJitter}} & \multirow{2}{*}{\textbf{mPQ}} \\ \cmidrule(lr){5-6}
 & & & & weak & strong &  \\ \midrule
Baseline                     & \chk &      & \chk &  & \chk & 50.71 \\
                             & \chk & \chk & \chk &  & \chk & 51.03 \\
                             &      & \chk & \chk &  & \chk & 51.07 \\ \midrule
HED-light                    &      & \chk &  &  &  & 50.48 \\ 
RandStainNA                  &      & \chk &  &  &  & 50.86 \\
RandStainNA$_{GMM}$          &      & \chk & & & & 50.71 \\ \midrule
\multirow{3}{*}{RandStainNA} &      & \chk & \chk &  &  &  51.07\\
                             &      & \chk & \chk & \chk &  & \underline{51.13}  \\
                             &      & \chk & \chk &  & \chk & 50.27 \\ \midrule
\multirow{3}{*}{RandStainNA$_{GMM}$} &      & \chk & \chk &  &  & 50.99 \\
                             &      & \chk & \chk & \chk &  & \textbf{51.61} \\
                             &      & \chk & \chk &  & \chk & 50.51  \\ \bottomrule
\end{tabular}
}
\vskip 1mm
\begin{minipage}{\linewidth}\raggedleft\footnotesize
where \textbf{S}: Solarization, \textbf{V}: Vertical Flip, \textbf{G}: Grayscale
\hspace*{0.3mm}
\end{minipage}
\vskip -1.5mm
\caption{\textbf{Augmentation ablation study on CoNSeP}.
mPQ scores are computed after fine-tuning with a 1$\times$ schedule. %
\textit{Baseline} refers to the original \bt setting.
} %
\vspace*{-4mm}
\label{result:bt_ablation}
\end{table}

\parahead{Augmentation Ablation}
We select nuclei instance segmentation as a target task, as it is one of the most practical and challenging tasks in pathology.
Starting from the default data augmentation of \bt \cite{zbontar2021barlow} denoted as \textit{Baseline} in~\autoref{result:bt_ablation}, we add a random vertical flip augmentation in order to take advantage of the nature wherein pathology images have no canonical orientation.
Based on prior work that claims that solarization can produce unrealistic and harmful images for pathology model training~\cite{faryna2021tailoring}, we exclude the random solarization augmentation.
With these two changes, we observe a gain of 0.36 in mPQ score.

We expect that stain augmentations serve to generate more domain-relevant augmented views, particularly, in terms of color variations.
However, stain augmentation clashes with color distortion, as both alter the color statistics of images.
Thus, we begin by disabling color distortion and then compare key stain augmentation methods first.
We find that RandStainNA~\cite{shen2022randstainna} and RandStainNA$_{GMM}$ out-perform HED-light~\cite{tellez2019quantifying}, confirming insights from supervised image classification~\cite{shen2022randstainna}.

Next, we bring back the components of color distortion (consisting of grayscale and color jitter) and evaluate them in detail.
We find that random grayscale augmentation is surprisingly effective, given that the produced images may not be considered plausible in pathology.
As the standard strength of color jittering produces highly unrealistic outputs, we evaluate a weaker jitter strength as well.
Indeed, we find that while the performance drops substantially when using strong color jitter, using weak color jitter together with random grayscale results in the best performances.
In particular, RandStainNA$_{GMM}$ shows high performance, motivating us to adopt it in our main experiments.

Through these observations, we substantiate our claim that existing augmentation schemes designed for SSL using natural images are sub-optimal for pathology data, necessitating pathology-specific augmentation schemes when training on pathology data such as TCGA and \lnt.

\begin{table}[]
\centering
\setlength{\tabcolsep}{3.0pt}
\begin{tabular}{lccccc}
\toprule
\multirow{2}{*}{FoV} & \multicolumn{4}{c}{\textbf{Top-1 Acc. (\%)}} & \textbf{mPQ} \\ \cmidrule(lr){2-5} \cmidrule(lr){6-6}
 & BACH & CRC & PCam & MHIST & CoNSeP \\ \midrule
20$\times$             &    {\ul 79.17} & \textbf{95.04} &    {\ul 85.24} &    {\ul 79.32} &    {\ul 50.66} \\
40$\times$             &    {\ul 79.17} &         91.88 &          80.82  &         74.21  &         48.83  \\
20$\times$, 40$\times$ & \textbf{85.00} &    {\ul 93.23} & \textbf{86.92} & \textbf{81.27} & \textbf{51.61}  \\
\bottomrule
\end{tabular}
\vskip -2mm
\caption{\textbf{Magnification ablation study.} Fine-tuning performance when using different FoVs during \pretrain ing.}
\vspace*{-4mm}
\label{result:nuclei-mag}
\end{table}

\parahead{Magnification Ablation}
In \autoref{method:technique_for_SSL}, we argue that pre-training using multiple magnifications or FoVs is beneficial as downstream pathology tasks occur at various magnifications.
We do find experimentally that using multiple FoVs in the pre-training dataset is beneficial for overall downstream task performance (see \autoref{result:nuclei-mag}).

Interestingly, we observe that using both 20$\times$ and 40$\times$ is best, while using only 20$\times$ is typically second-best.
This is the case even for datasets such as \pcam, MHIST, and CoNSeP which consist of images collected at approximately 40$\times$.
We hypothesize that the use of multiple magnifications is not valuable due to the matching of magnifications between upstream and downstream training, but rather due to the diversity of image appearances.
Specifically, 20$\times$ images, due to the wider field-of-view, are visually and texture-wise more diverse than 40$\times$ images.
Combining the two magnifications results in an even more diverse set of images.
The more diverse data also results in better convergence during \pretrain ing (see supplementary materials).

\section{Discussion}

In this section, we answer a few key questions that computational pathology researchers may naturally ask when considering self-supervised \pretrain ing for their research.

\parahead*{Should we \pretrain~on pathology data?} Yes -- We have consistently demonstrated that pre-training on pathology data out-performs supervised pre-training on ImageNet by performing comprehensive experiments on many SSL methods and datasets.
Interestingly, SSL pre-trained weights can maintain the performance gap on CoNSeP even for longer training schedules.
Our experiments demystify and confirm the effectiveness of domain-aligned SSL \pretrain ing on the pathology domain.

\parahead*{Which SSL method is best?} We find that there is \emph{no clear winner}. All SSL methods applied with domain-aligned \pretrain ing generally perform well.
Thus, instead of focusing on selecting a specific SSL method, we recommend that practitioners focus on curating large-scale domain-aligned datasets for SSL \pretrain ing.
Yet, some initial observations may be useful to future research. For example, (a) \bt tends to perform well in linear evaluations and \moco in fine-tuning evaluations, and (b) ViTs benefit more from domain-aligned SSL compared to CNNs.

\parahead*{What is a key ingredient for successful self-supervised \pretrain ing?}
Domain knowledge -- %
our proposed set of techniques are fully based on observations in pathology, and are experimentally shown to be effective.
By incorporating domain-specific knowledge into the \pretrain ing step, e.g., using stain augmentation and extracting patches from multiple FoVs, we go beyond the performance one can get from naively applying SSL to a new dataset.

\section{Conclusion and Future Work}
In this paper, we conduct the largest and most comprehensive study of SSL in the pathology domain, to date, using up to 33 million image patches during pre-training and evaluating 4 representative SSL methods (both CNNs and ViTs) on 2 downstream tasks and 5 datasets. 
Our study confirms that large-scale domain-aligned \pretrain ing is helpful for pathology, showing its value in scenarios with limited labeled data, longer fine-tuning schedules, and when using larger and more diverse datasets for \pretrain ing (such as TCGA + \lnt).
Furthermore, we propose a set of techniques that are carefully designed by leveraging pathology-specific knowledge, and integrate them into the self-supervised \pretrain ing stage, resulting in performance improvements.
We believe that further exploration of domain-specific augmentation strategies will yield improved techniques for pathology-specific SSL in the future.

{\small
\bibliographystyle{ieee_fullname}
\bibliography{egbib}
}

\appendix
\counterwithin{figure}{section}
\counterwithin{table}{section}
\section*{Appendix}

\noindent\textbf{Overview.}\quad In this supplementary material, we describe the details of the downstream datasets adopted in the main paper and show some example images. This document also contains further implementation details regarding the \pretrain ing and downstream training steps, including fine-tuning with limited labeled data. Last but not least, we provide further analyses, such as the effectiveness of \pretrain ing for longer epochs and \pretrain ing stability when using data from different magnifications.

Note that the corresponding or relevant sections from the main paper are referenced 
Note that the corresponding or relevant sections from the main paper are referenced \textbf{\textcolor{blue}{in blue text}} in the section titles.

\section{Downstream Dataset Details \color{blue}{(Section 4.2)}}

\begin{figure}[t]
    \centering
    \vskip -1mm
    \begin{tabular}{@{}m{0.07\columnwidth}@{}m{0.46\columnwidth}@{}m{0.46\columnwidth}@{}}
        (a) & \includegraphics[width=0.45\columnwidth]{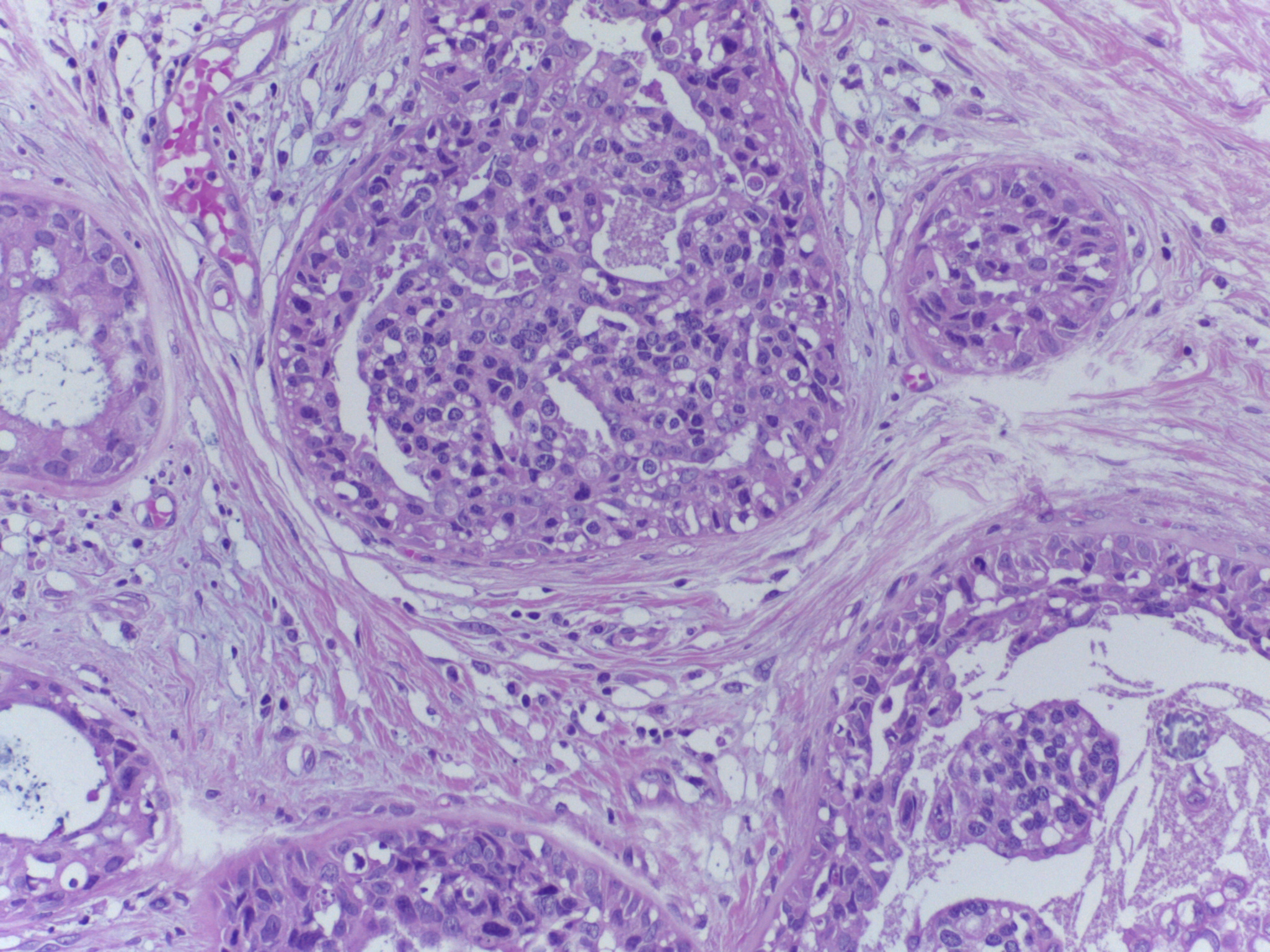} & \includegraphics[width=0.45\columnwidth]{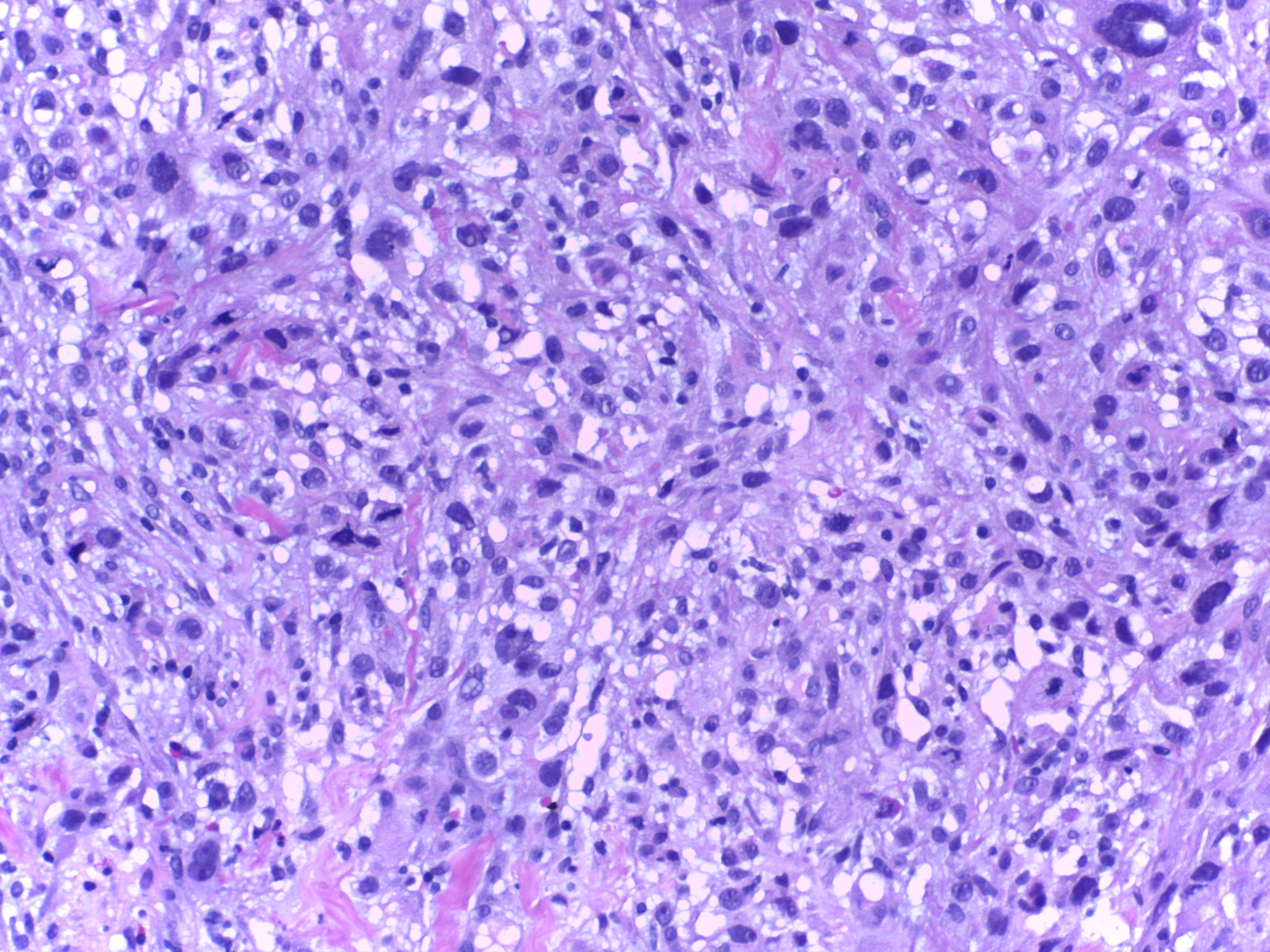} \\
        (b) & \includegraphics[width=0.45\columnwidth]{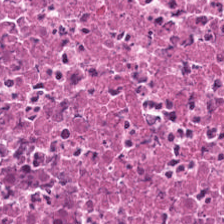} & \includegraphics[width=0.45\columnwidth]{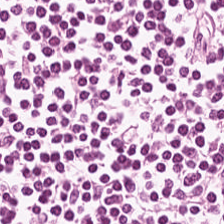} \\
        (c) & \includegraphics[width=0.45\columnwidth]{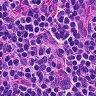} & \includegraphics[width=0.45\columnwidth]{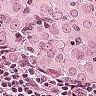} \\
        (d) & \includegraphics[width=0.45\columnwidth]{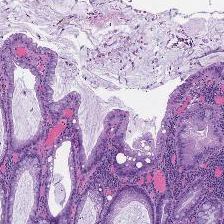} & \includegraphics[width=0.45\columnwidth]{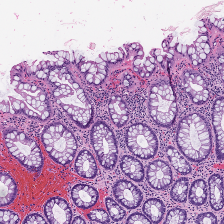} \\
    \end{tabular}
    \vskip -3mm
    \caption{\textbf{Example training images from the classification datasets}: (a) BACH, (b) CRC, (c) PCam, and (d) MHIST.%
    }
    \vskip -2mm
    \label{fig:classification_example}
\end{figure}

\begin{figure}
    \centering
    \includegraphics[width=\linewidth]{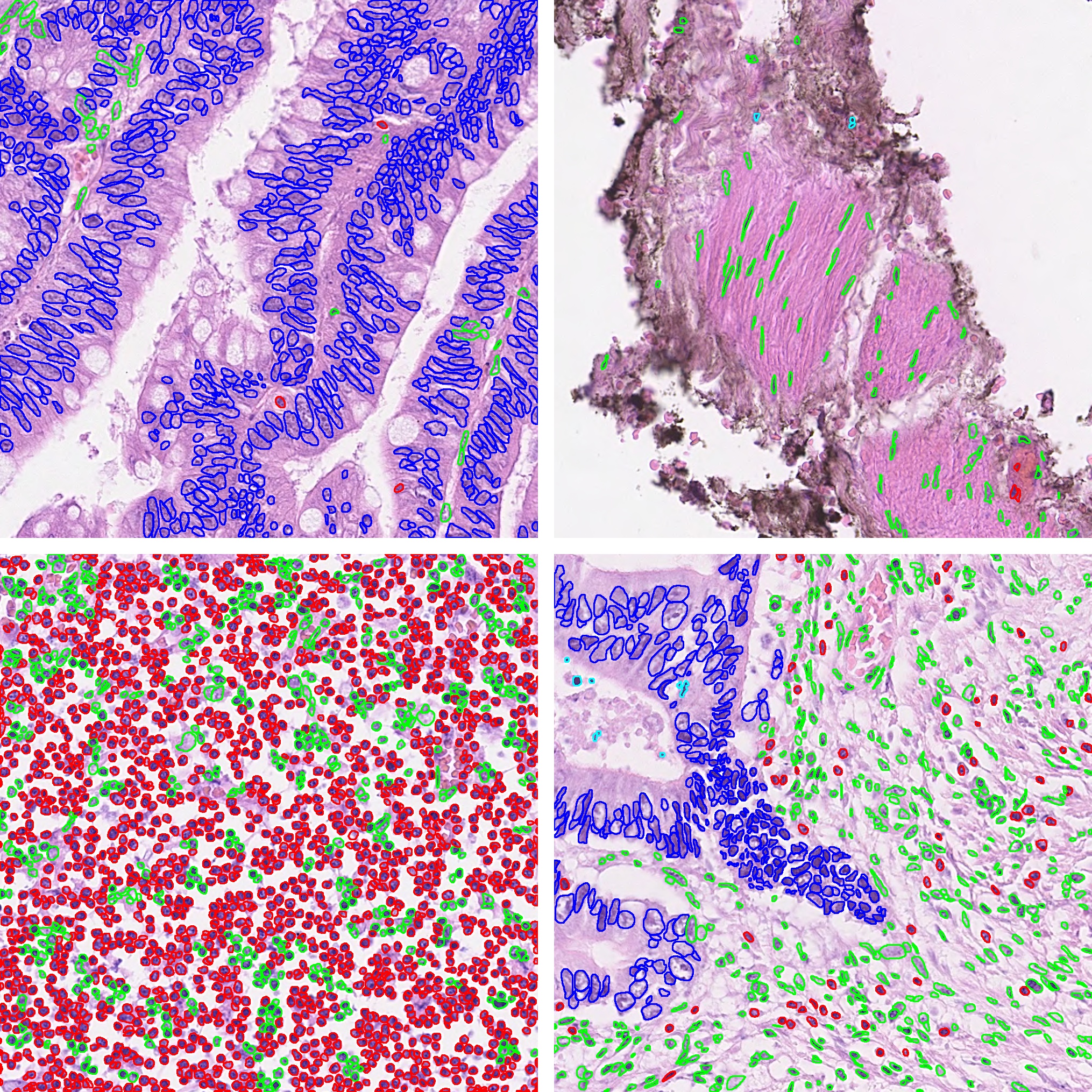}
    \vskip -2mm
    \caption{\textbf{Example training images from the CoNSeP dataset.} The dataset provides annotated nuclei masks along with cell type labels. Following the original HoVer-Net paper~\cite{graham2019hover}, we use the following nuclei types for training and evaluation: 
    {\textcolor{blue}{$\mdlgblksquare$}}~epithelial,
    {\textcolor{red}{$\mdlgblksquare$}}~inflammatory,
    {\textcolor{green}{$\mdlgblksquare$}}~spindle-shaped, and
    {\textcolor{cyan}{$\mdlgblksquare$}}~miscellaneous.
    }
    \label{suppl:consep_example}
\end{figure}

In this section, we describe the details of the datasets used in our analysis. 
We use BACH, CRC, PCam, and MHIST for the image classification task, and CoNSeP for the nuclei instance segmentation task.
We sample a few training images from each dataset and present them in~\autoref{fig:classification_example} and ~\autoref{suppl:consep_example}.
\textbf{\textcolor{blue}{in blue text}} in the section titles.

\parahead{BACH}
The goal of the Grand Challenge on BreAst Cancer Histology (BACH)~\cite{aresta2019bach} is to classify pathology images into four classes: normal, benign, in situ carcinoma, and invasive carcinoma. The dataset is composed of 400 training images and 100 test images. The test images are collected from a different set of patients from the training images. All images are collected from Hospital CUF Porto, Centro Hospitalar do Tâmega e Sousa, and Centro Hospitalar Cova da Beira. \par

\parahead{CRC}
This dataset~\cite{kather_jakob_nikolas_2018_1214456} consists of 100,000 training images and 7,180 test images from H\&E stained WSIs of human colorectal cancer (CRC) and normal tissue. The training and test images are extracted from 86 WSIs and 25 WSIs, respectively. The slides are collected from the NCT Tissue Bank and the University Medical Center Mannheim. The task is the identification of nine tissue classes: adipose tissue, background, debris, lymphocytes, mucus, smooth muscle, normal colon mucosa, cancer-associated stroma, and CRC epithelium. 
All images are color normalized with the Macenko method~\cite{macenko2009method}. 

\parahead{PCam}
The PatchCamelyon (PCam)~\cite{veeling2018rotation} dataset is derived from the Camelyon16~\cite{bejnordi2017diagnostic} dataset that contains 400 H\&E stained WSIs from two hospitals: Radboud University Medical Center (RUMC), and University Medical Center Utrecht (UMCU). The PCam dataset includes 262,144 training images, 32,768 validation images, and 32,768 test images. Each image is annotated with a binary label for determining the presence of metastases. \par

\parahead{MHIST}
The minimalist histopathology image analysis (MHIST)~\cite{wei2021petri} dataset is comprised of 2,175 training images and 977 test images. The images are extracted from 328 H\&E stained Formalin Fixed Paraffin-Embedded (FFPE) WSIs of colorectal polyps from Dartmouth-Hitchcock Medical Center. The task is the binary classification between hyperplastic polyps (HPs) and sessile serrated adenomas (SSAs), where HPs are benign and SSAs are precancerous lesions. \par

\parahead{CoNSeP}
The Colorectal Nuclear Segmentation and Phenotypes (CoNSeP) dataset \cite{graham2019hover} consists of 41 H\&E images and is split into 27 images and 14 images for training and test sets, respectively.
The data comes from University Hospitals Coventry and Warwickshire, UK (UHCW). %
The annotation contains segmentation masks of each nucleus along with its class (See~\autoref{suppl:consep_example}).
Note that the healthy epithelial and dysplastic/malignant epithelial are considered general epithelial types.
Fibroblast, muscle, and endothelial are matched into a spindle-shaped nuclei type.
In total, 24,319 unique nuclei masks along with 4 major types out of 7 cell types are used during training.

\section{Implementation Details}

In the interest of improving the reproducibility of our study, we provide further details regarding our \pretrain ing data, setup, as well as details on how we conducted our downstream evaluations. Furthermore, we discuss the details of the limited labeled data experiments.

\subsection{Preparation of Pre-training Data \color{blue}{(Section 4.1)}} 

In selecting image patches to compose the TCGA dataset, we first use an internal model with a DeepLab v3+ architecture~\cite{chen2018encoder} to segment the foreground regions of WSI.
From the candidate patches that are located in areas predicted as foreground, we select up to 500 patches per magnification, per slide, with equal spacing between them.
To ensure that we have informative image patches in our pre-training dataset, we filter out patches that are too white (mean saturation value below 5) or too smooth (mean squared Laplacian below 15).
For \lnt, we do not apply such filtering logic due to the relatively smaller foreground area (too many patches are lost otherwise).

\subsection{Calculating Statistics of the Pre-training Data \color{blue}{(Section 4.1)}}
For the purpose of input image standardization during SSL pre-training, we collect the per-channel mean and standard deviation of intensities in RGB space, using 10\% of the full unlabeled image data. This subsampling is done per WSI, to maintain diversity and reduce computational cost.

In a similar manner, we compute the per-channel means and variances in 3 color spaces (HSV, Lab, HED) for use with the RandStainNA method, using 10\% of the full image data.
Specifically, we compute per color space, and per channel, the mean and standard deviation of per-image mean intensity, as well as the mean and standard deviation of the per-image standard deviation of intensity.
Please refer to \cite{shen2022randstainna} for further details.

For RandStainNA$_{GMM}$, we similarly compute per color space, and per channel, the per-image mean intensity and its standard deviation.
However, instead of simply finding the mean and standard deviation of those values independently (fitting individual unimodal Gaussian distributions 18 times as in RandStainNA), we fit a 10-component Gaussian Mixture Model (GMM) for each color space, yielding 3 models.
This is done to fit the covariance between the input variables (6 variables exist for each color space) and respect their multi-modal nature.

\subsection{Augmentation Details \color{blue}{(Section 3.2)}}
Unless otherwise stated, in our experiments, we \pretrain~by applying the following changes to the default method-specific augmentation scheme:
\begin{itemize}
    \item Random vertical flip \texttt{(p=0.5)}.
    \item Color dropping \texttt{(p=0.2)}: the color of images are converted randomly to grayscale.
    \item Weak color jittering \texttt{(p=0.8)}: the brightness, contrast, saturation, and hue of images are randomly adjusted with a strength of \texttt{0.2, 0.2, 0.2, 0.1}, respectively.
    \item RandStainNA$_{GMM}$ \texttt{(p=0.8)}: per image, a color space is randomly selected (from HSV, Lab, or HED), then channel-wise mean and standard deviation values are sampled from a GMM which is fitted on statistics from part of the \pretrain ing data (10\%). The input image is re-normalized based on these values, using Reinhard's method~\cite{reinhard2001color}.
\end{itemize}

\subsection{SSL Methods \color{blue}{(Section 4.3)}}

We provide implementation details of each SSL method used in our analysis.
We use the VISSL \cite{goyal2021vissl} library to \pretrain~the the 4 studied SSL methods, and follow the same configurations as originally proposed in \cite{chen2020improved,zbontar2021barlow,caron2020unsupervised,caron2021emerging}. %
All representations are trained for 200 ImageNet epochs, distributed over 64 V100 16GB GPUs.
A linear warmup schedule is applied for the first 10 epochs and a cosine learning rate decay is applied subsequently.
Each method was originally proposed with its specific augmentation schemes, and we follow those original data augmentation pipelines while adding our proposed techniques on top.
Regarding the RandStainNA augmentation, it requires the statistics of 3 color spaces (HSV, Lab, HED) to produce augmented images.
To compute the statistics, we randomly sample 10\% of the unlabeled image patches from the corresponding \pretrain ing dataset.

\parahead{\moco} 
We use the SGD optimizer with an initial learning rate of 0.3.
The learning rate is linearly scaled up based on \textit{lr} = $lr * batchsize$/256, where \textit{batchsize} is 4,096.
The memory bank size is fixed to 65,536, and a momentum coefficient \textit{m} of 0.999 is used.
Weight decay of $ 10^{-4}$ is utilized for regularization.

\parahead{SwAV}
We use the SGD optimizer with an initial learning rate of 0.3.
The learning rate is linearly scaled up based on \textit{lr} = $lr * batchsize$/256, where \textit{batchsize} is 2,048.
The number of prototypes is 3,000 to avoid intractable computational costs from the Sinkhorn algorithm.
2$\times$224 + 6$\times$96 multi-crop augmentation is employed as done in the original paper.

\parahead{\bt}
The LARS optimizer~\cite{you2017large} is adopted for \bt \pretrain ing.
Note that, as in the original work \cite{zbontar2021barlow}, we apply different learning rates for weights and biases, 0.2 and 0.0048, respectively.
The biases and batch normalization layers are excluded from LARS optimization to follow the original implementation. 
The learning rates of weights and biases are linearly scaled up based on \textit{lr} = $lr * batchsize$/256, where \textit{batchsize} is 2,048.
The dimension of the embeddings is 8,192, and training is conducted with a coefficient of off-diagonal term $\lambda=5\cdot10^{-3}$ and a weight decay of $1.5 \cdot 10^{-6}$. %

\parahead{DINO}
We train the model with the AdamW~\cite{loshchilov2018fixing} optimizer.
The learning rate of $0.0005$ is used for stability during \pretrain ing.
The learning rate is linearly scaled up based on \textit{lr} = $lr * batchsize$/256, where \textit{batchsize} is 1,024.
Similar to the learning rate decay, the weight decay also follows a cosine schedule from 0.04 to 0.4. 
For DINO$_{p=16}$, 2$\times$ 224 + 8$\times$ 96 multi-crop augmentation is employed, while 2$\times$ 224 + 6$\times$ 96 multi-crop augmentation is used for DINO$_{p=8}$.%

\begin{table*}[th]
\centering
\setlength{\tabcolsep}{3pt}
\renewcommand{\arraystretch}{0.9}
\begin{tabular}{@{}lllccccccc@{}}
\toprule
\multirow{2}{*}{\textbf{Arch.}} & \multirow{2}{*}{\textbf{Method}} & \multicolumn{2}{c}{\textbf{BACH}} & \multicolumn{2}{c}{\textbf{CRC}} & \multicolumn{2}{c}{\textbf{\pcam}} & \multicolumn{2}{c}{\textbf{MHIST}} %
\\
&                                  & Linear          & Fine-tune       & Linear          & Fine-tune      & Linear           & Fine-tune       & Linear           & Fine-tune       %
\\ \midrule
\multirow{10}{*}{ResNet-50}
& \textit{Random}                  & 51.67           & 61.67           & 68.91           & 89.99          & 76.52            & 75.71           & 63.15            & 75.54 %
\\
& \textit{Supervised}              & 80.83           & 86.67     & 90.93           & 92.09          & 80.79            & 80.63           & 76.25            & 78.92 %
\\ \cmidrule{2-10}
& {\footnotesize \ul \textbf{Epoch 200}}    &                   &                   &                   &                   &                   &                  \\[1mm]
& \moco                            & 77.50           & {\ul 90.83}  & 93.52           & \textbf{96.21} & 86.78     & \textbf{87.62}  & 77.07      & \textbf{85.88} %
\\
& SwAV                             & {\ul 83.33}     & 82.50           & {\ul 95.78}  & 93.31    & 85.28            & {\ul 87.60}     & 71.14            & 77.99 %
\\
& BT                               & \textbf{87.50}  & 85.00           & 94.60     & 93.23          & \textbf{88.15}   & 86.92           & \textbf{78.81}   & 81.27 %
\\
\cmidrule(l){2-10}
& {\footnotesize \ul \textbf{Epoch 800}}    &                   &                   &                   &                   &                   &                  \\[1mm]
& \moco                             & 79.17         & \textbf{91.67}        &  95.01           & {\ul 95.45}        & {\ul 87.84}          & 86.90       & 72.77            & 84.95 %
\\
& SwAV                             & 82.50             & 85.83           & \textbf{96.46}            & 92.74        & 86.16          & 87.05       & 75.54            & {\ul 85.47} %
\\
& BT                               & 86.67             & \textbf{91.67}            & 94.48            & 94.99        & 86.26          & 86.75       & {\ul 78.20}            & 80.25 %
\\
\midrule
\multirow{8}{*}{ViT-S}
& \textit{Random$_{p=16}$}         & 45.00           & 57.50           & 69.90           & 86.10          & 74.43            & 75.42           & 63.46            & 62.13  %
\\
& \textit{Supervised$_{p=16}$}     & 75.83           & 85.83           & 91.56           & {\ul 95.81}    & 80.96            & 88.30           & \textbf{78.51}   & \textbf{81.68} %
\\ \cmidrule{2-10}
& {\footnotesize \ul \textbf{Epoch 200}}    &                   &                   &                   &                   &                   &                  \\[1mm]
& DINO$_{p=16}$                    & {\ul 85.83}  & {\ul 87.50}     & {\ul 94.19}     & {\ul 95.81}    & \textbf{88.78}      & {\ul 90.40}     & {\ul 76.15}            & 79.43 %
\\
\cmidrule(l){2-10}
& {\footnotesize \ul \textbf{Epoch 400}}    &                   &                   &                   &                   &                   &                  \\[1mm]
& DINO$_{p=16}$                    &   \textbf{86.67}           &   \textbf{88.33}          &   \textbf{95.13}          & \textbf{96.48}        & {\ul 88.60}          &  {\ul 89.50}      &   75.44          &  {\ul 81.06}         \\
\bottomrule
\end{tabular}
\caption{\textbf{Downstream evaluation of image classification tasks under a different number of \pretrain ing epochs.} We report Top-1 accuracy for both linear and fine-tuning experiment protocols trained using the TCGA data source. Note that $p$ represents the patch size used in ViT.
We compare results column-wise and mark the best results in \textbf{bold} and the second-best results in \underline{underline} for ResNet-50 based methods and ViT-S methods separately.
}
\label{suppl:pretraining_epochs}
\end{table*}

\subsection{Downstream Training Details \color{blue}{(Section 4.4)}}
\parahead{Image Classification}
We split each downstream dataset into training, validation, and test sets.
The learning rate and weight decay values are optimized using training and validation, only. 
In the BACH dataset, the labels for the test set are not provided. Hence, we split the training set by a 6:1:3 ratio (training, validation, test). For the CRC and MHIST datasets, the test set is provided with labels, and the training set is split by a 7:3 ratio (training, validation). For the PCam dataset, we follow the original data split.
When splitting the data, we do it randomly but in a class-balanced manner.
Based on the performance measured on the validation sets, we perform a grid search of learning rates from $\{1, 0.1, 0.01, 0.001\}$ and weight decay values from $\{0.1, 0.01, 0.001, 0\}$.

As data augmentation for ResNet-50, the input image is randomly flipped both horizontally and vertically, at training time. For the BACH dataset, we apply random cropping and resizing to $1024 \times 768$ at training time; at test time, we resize the images to $1024 \times 768$. For ViT-S, the same augmentation is used but all images are resized to $224 \times 224$.
We train the models with the SGD optimizer with a momentum of 0.9 and a cosine learning rate decay. The ResNet-50 based models are trained for 200, 20, 20, and 90 epochs on the BACH, CRC, PCam, and MHIST datasets, respectively. 
The Transformer-based models are trained for 30 epochs on the CRC and PCam datasets and for 200 and 90 epochs on the BACH and MHIST datasets, respectively. During fine-tuning, the backbone layers (i.e., ResNet-50 and ViT-S) are trained with a learning rate 100 times lower than that of the last classification layer.

\parahead{Nuclei Instance Segmentation}
We follow the standard pipeline of HoVer-Net \cite{graham2019hover}, as provided in its open-source implementation\footnote{\url{https://github.com/vqdang/hover_net}}, including data augmentation and patch extraction.
Hover-Net defines a two-stage training procedure.
At the first stage, only the decoders are trained while freezing the backbone layers. 
With the trained decoders, all layers are then fine-tuned at the second stage.
Technically, Preact-ResNet-50~\cite{he2016identity} is employed in the original implementation, but we replace it with the standard ResNet-50 \cite{he2016deep} and reproduce the results for a fair comparison. 
This is done to perform SSL \pretrain ing in a standard manner while permitting this nuclei instance segmentation downstream task.
Since we change the backbone, we perform Grid Search to find a proper learning rate. 
We use $5\cdot10^{-4}$ learning rate for both stages of HoVer-Net.
Moreover, based on the open-source implementation, the authors of HoVer-Net fine-tune the first convolutional layer of ResNet at the first stage, but we keep them frozen.

Similar to the architecture of FPN-based instance segmentation, HoVer-Net requires features from multiple scales in the encoder.
However, the outputs of the ViT-based encoder are not compatible with the existing decoders of Hover-Net without further modifications.
In order to provide multi-scale features to the decoder, we refer to the protocol from~\cite{ali2021xcit} where the feature scales are interpolated using several transposed convolution layers with kernel size $k=2$ and stride $s=2$.
More specifically, features from the $4^{th}$, $6^{th}$, $8^{th}$, and $12^{th}$ layers are extracted from the ViT-S architecture, which consists of 12 layers in total. %
With this design, the decoders remain unchanged.
For the sake of a fair comparison, we also perform Grid Search on the ViT-S architecture.
The learning rate of $5\cdot10^{-4}$ is used for both stages of HoVer-Net.

\subsection{Fine-tuning with Limited Labeled Data \color{blue}{(Section 5.4)}}

\parahead{Image Classification}
Following prior works~\cite{chen2020simple, grill2020bootstrap, zbontar2021barlow}, we randomly sample 1\% and 10\% of the CRC training set by balancing classes. Based on the fine-tuning procedure, we train the models for 60 and 90 epochs for 1\% and 10\% labeled data, respectively.

\parahead{Nuclei Instance Segmentation using CoNSeP}
In our limited labeled data experiments using CoNSeP, we control the number of H\&E images instead of the number of extracted patches to mimic the real-world setting where one H\&E image corresponds to one unique patient.
Since assuming 1\% of training data is unreasonable in the current setting (i.e, 0.27 H\&E image), instead, we define the ratio of 10\% and 30\% for nuclei instance segmentation.
Note that the reported values in the experiments are the averaged number from 3 repetitive experiments with different seed values for image/patient selection.
This is necessitated by the smaller dataset size (compared to CRC) and is done for a fair comparison between methods.

\section{Pre-training for more epochs \color{blue}{(Section 5)}}

\begin{table}[t]
\centering
\setlength{\tabcolsep}{3pt}
\renewcommand{\arraystretch}{0.80}
\begin{tabular}{@{}llcc@{}}
\toprule
\multirow{2}{*}{\textbf{Arch.}} & \multirow{2}{*}{\textbf{Method}} & \multicolumn{2}{c}{\textbf{CoNSeP}}   \\
                                &                                  & Linear            & Fine-tune         \\ \midrule
\multirow{10}{*}{ResNet-50}      & \textit{Random}                  & 22.29             & 46.72             \\
                                & \textit{Supervised}              & 34.25             & 49.60             \\ \cmidrule{2-4}
                                & {\footnotesize \ul \textbf{Epoch 200}}    &                   &    \\[1mm]
                                & \moco                            & 39.85             & 51.75    \\
                                & SwAV                             & 40.45 & 51.16             \\
                                & BT                               & \underline{40.79}    & 51.61 \\ \cmidrule{2-4}
                                & {\footnotesize \ul \textbf{Epoch 800}}    &                   &    \\[1mm]
                                & \moco                            & \textbf{40.93} & 51.64    \\
                                & SwAV                             & 40.59 & \textbf{52.39}             \\
                                & BT                               & \underline{40.65} & \underline{52.00} \\ \midrule
\multirow{8}{*}{ViT-S}          & \textit{Random$_{p=16}$}         & 20.55             & 27.19             \\
                                & \textit{Supervised$_{p=16}$}     & 21.43             & 36.70              \\ \cmidrule{2-4}
                                & {\footnotesize \ul \textbf{Epoch 200}}    &                   &    \\[1mm]
                                & DINO$_{p=16}$                    & {\ul 32.54} & {\ul 38.43} \\ \cmidrule{2-4}
                                & {\footnotesize \ul \textbf{Epoch 400}}    &                   &    \\[1mm]
                                & DINO$_{p=16}$                    & \textbf{32.93} & \textbf{39.03}  \\
                                \bottomrule
\end{tabular}
\caption{\textbf{Downstream evaluation for the nuclei instance segmentation task under a different number of \pretrain ing epochs}. We report the mPQ score for both linear and fine-tuning experiment protocols for models trained using the TCGA data source. 
We compare results column-wise and mark the best results in \textbf{bold} and the second-best results in \underline{underline} for ResNet-50 based methods and ViT-S methods separately.
}
\label{suppl:qualitative_segmentation _results}
\end{table}

\begin{figure*}[t]
    \centering
    \includegraphics[width=\linewidth]{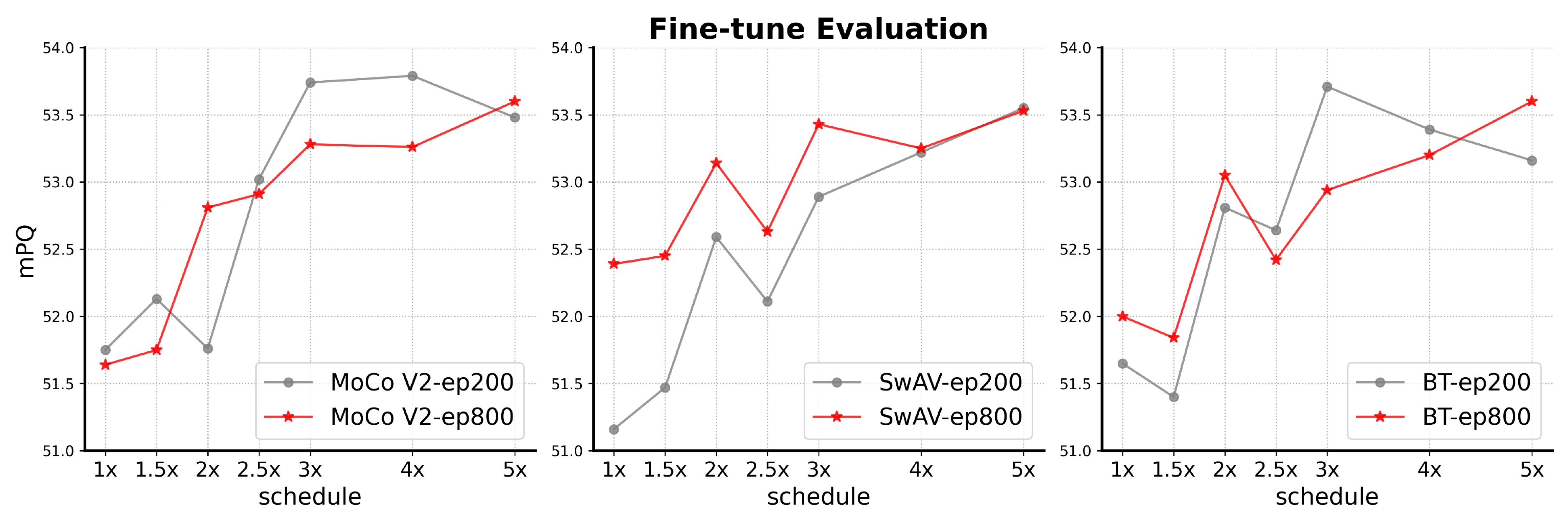}
    \caption{\textbf{The effectiveness of longer \pretrain ing according to learning schedules.} We present 
    fine-tuning evaluation results for the nuclei instance segmentation task using the CoNSeP dataset.
    We see that there are few differences between the 200 epoch models and 800 epoch models, except that SwAV benefits from longer \pretrain ing when the downstream task is fine-tuned with a limited learning schedule.
    }
    \label{suppl:nuc_sch_study}
\end{figure*}

Typically, increasing the number of \pretrain ing epochs has shown to be effective in improving the learned representations in various SSL methods.
To investigate the effectiveness of the longer \pretrain ing in the pathology domain, we pre-train the model for 800 ImageNet epochs using \moco, SwAV, and \bt. 
Note that, due to computational costs, we report the results from DINO$_{p=16}$ trained for 400 ImageNet epochs.

\autoref{suppl:pretraining_epochs} and \autoref{suppl:qualitative_segmentation _results} present the performance of image classification and nuclei instance segmentation, respectively.
Compared to the results from 200 ImageNet epochs, SwAV is the only method that benefits from the longer \pretrain ing in the fine-tuning protocol, especially in BACH, MHIST, and CoNSeP datasets. In contrast, the other methods show marginal improvements or are on par with the 200 ImageNet epoch counterparts.
DINO$_{p=16}$ shows a slightly improved performance on image classification, while nuclei instance segmentation remains on par. %
Even in the different learning schedules illustrated in~\autoref{suppl:nuc_sch_study}, we observe that no clear benefit of the longer \pretrain ing stands out in \moco and \bt, yet SwAV consistently maintains the benefit of the longer \pretrain ing.

Across all experiments, we confirm that certain SSL methods (e.g., SwAV) may require more \pretrain ing iterations, but generally increasing the number of \pretrain ing epochs shows marginal improvements on both image classification and nuclei instance segmentation tasks.
In other words, \pretrain ing for 200 ImageNet epochs can be sufficient to achieve satisfactory downstream performance, especially for \moco, \bt, and DINO.
We therefore suggest that using 200 ImageNet epochs would be adequate to study the potential of SSL \pretrain ing in the pathology domain.

\section{Pre-training Stability with Different Magnifications \color{blue}{(Section 5.6)}}

\begin{figure}
    \centering
    \includegraphics[width=\linewidth]{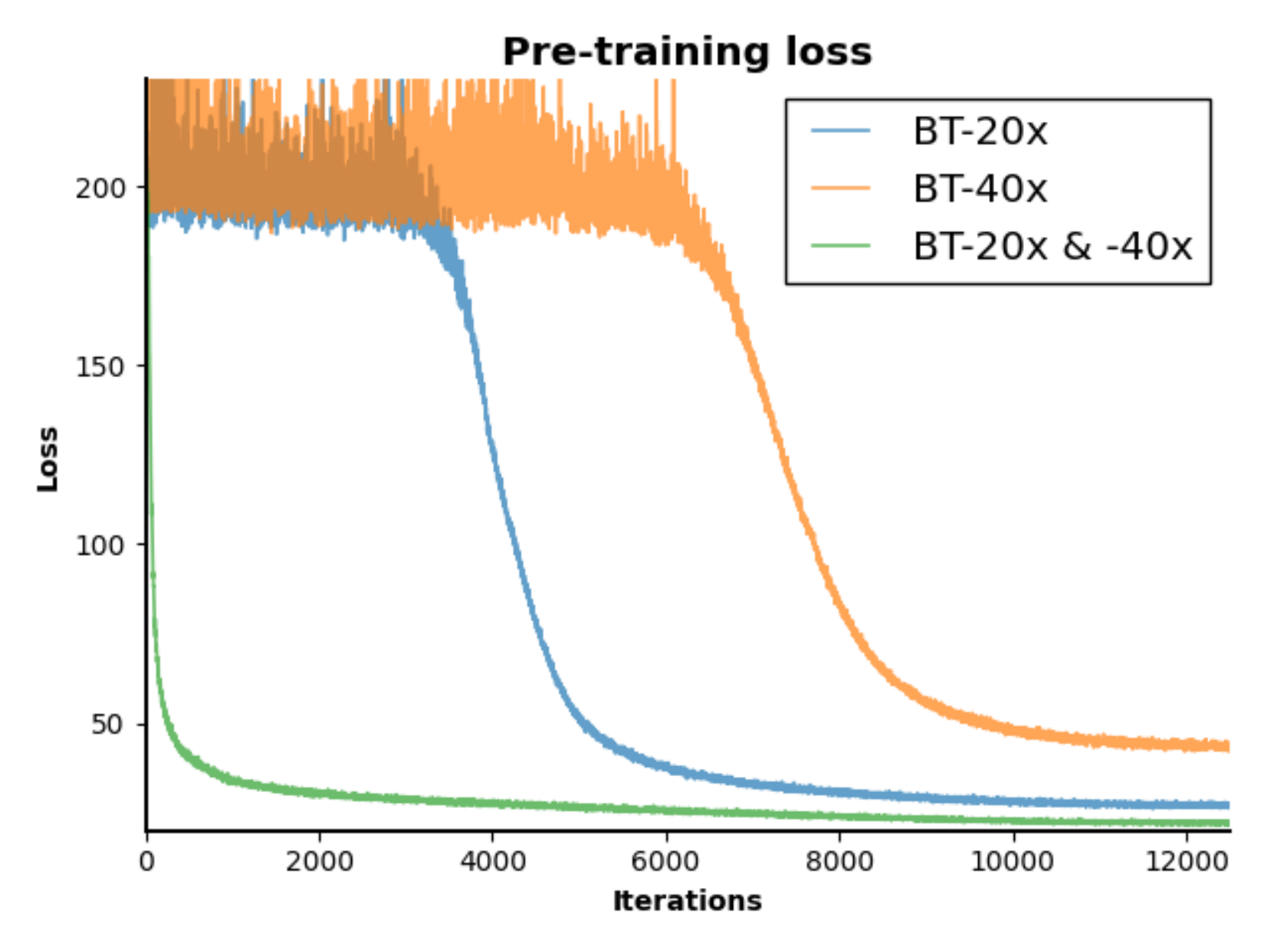}
    \caption{\textbf{Loss progression while \pretrain ing \bt on different magnifications.}
    Training on a combination of $20\times$ and $40\times$ results in quick convergence and stable \pretrain ing.}
    \label{suppl:bt_fov_study}
\end{figure}

In the main paper, we show that it is beneficial to train on image data from a combination of $20\times$ and $40\times$ objective magnifications.
Here, we show that \pretrain ing stability is also affected by the choice of magnification.
In \autoref{suppl:bt_fov_study}, we present the loss trajectory during the \pretrain ing stage using \bt.
As shown in the graph, using a single magnification produces unstable losses and the loss begins to converge after approximately 4,000 and 7,000 iterations for magnifications of 20$\times$ and 40$\times$, respectively.
The loss values at the end of the \pretrain ing stage are also higher in the case of using a single magnification.
In contrast, using multiple magnifications results in stable \pretrain ing and fast convergence, in addition to improved downstream task performance.

\section{Larger Inputs for ViT \color{blue}{(Section 5.2)}}

The implementation of the standard HoVer-Net \cite{graham2019hover} method involves the fine-tuning of a pre-trained ResNet, using images with a resolution of 270 $\times$ 270. 
However, by design, ViT expects input images of 224 $\times$ 224 resolution.
Given the potential advantages that larger input resolutions can bring to the task of nuclei instance segmentation, we adopt a positional embedding interpolation technique to increase the input image size to 272 $\times$ 272, which is divisible by both 16 and 8. 
Through this technique, we aim to maintain consistent input resolutions across the ResNet and ViT backbones being evaluated.
\autoref{suppl:larger_input} presents the result according to the input size.
We observe that the larger input size improves performance for DINO$_{p=16}$, while the performance of DINO$_{p=8}$ reduces.

\begin{table}[t]
\centering
\setlength{\tabcolsep}{3pt}
\renewcommand{\arraystretch}{0.80}
\begin{tabular}{@{}llcc@{}}
\toprule
\multirow{2}{*}{\textbf{Arch.}} & \multirow{2}{*}{\textbf{Method}} & \multicolumn{2}{c}{\textbf{CoNSeP}}   \\
                                &                                  & Linear            & Fine-tune         \\ \midrule
                                & {\footnotesize \ul \textbf{224 input}}    &                   &    \\[1mm]
\multirow{7}{*}{ViT-S}          & \textit{Supervised$_{p=16}$}     & 21.43             & 36.70              \\  
                                & DINO$_{p=16}$                    & {\ul 32.54} & {\ul 38.43} \\  
                                & DINO$_{p=8}$                     & \textbf{42.71}    & \textbf{46.70}     \\   \cmidrule{2-4}
                                & {\footnotesize \ul \textbf{272 input}}    &                   &    \\[1mm]
                                & \textit{Supervised$_{p=16}$}     & 28.60      &  34.50            \\  
                                & DINO$_{p=16}$                    & {\ul 35.81} &  {\ul 41.13}  \\
                                & DINO$_{p=8}$                     & \textbf{40.08}    & \textbf{44.24} \\ 
                                \bottomrule
\end{tabular}
\caption{\textbf{Downstream evaluation for the nuclei instance segmentation task under a different input resolution.} We report the mPQ score for both linear and fine-tuning experiment protocols for models trained using the TCGA data source. We compare results column-wise and mark the best results in \textbf{bold} and the second-best results in {\ul underline}.
}
\label{suppl:larger_input}
\end{table}

\section{Further Data Augmentation Ablation Study \color{blue}{(Section 5.6)}}

\begin{table}[t]
    \centering
    \setlength{\tabcolsep}{1pt}
    \renewcommand{\arraystretch}{0.8}
    \begin{tabular}{@{}lccccc@{}}\toprule
         & BACH & CRC & PCam & MHIST & CoNSeP \\ \midrule
         BT trained on TCGA
         & 84.2 & 94.2 & 84.5 & 78.0 & 40.9 \\
         + our aug. techniques
         & \textbf{87.5} & \textbf{94.7} & \textbf{87.6} & \textbf{79.5} & \textbf{41.3} \\
         \bottomrule
    \end{tabular}
    \caption{ 
    \textbf{Benefit of our augmentation techniques.}
    Linear evaluation results show that our proposed augmentation techniques consistently and significantly improve performance.
}
    \label{ablation:techniques}
    \vspace*{-3mm}
\end{table}

To provide a compelling demonstration of the effectiveness of the proposed techniques, we opted for the most practical, yet challenging fine-tuning setting of nuclei instance segmentation. %
Through the application of the linear evaluation protocol, we further validate the effectiveness of our techniques by showcasing improvements across all datasets. 
Notably, our set of techniques consistently and significantly improves the performance compared to the baseline approach that relies on augmentations designed for natural images.
The improvement presented in \autoref{ablation:techniques} serves as a clear signal of the effectiveness of our proposed techniques, which were carefully designed with the aid of domain-specific knowledge. %

\section{Intriguing Properties of Self-supervised ViT \color{blue}{(Section 5.2)}}
\begin{figure*}[t]
\center
\vspace{2mm}
\renewcommand{\tabcolsep}{0.9mm}
\begin{tabular}{lcccccc}
     & \textbf{Head \#1} & \textbf{Head \#2} & \textbf{Head \#3} & \textbf{Head \#4} & \textbf{Head \#5} & \textbf{Head \#6} \\ \vspace{-5.0mm}
  \rotatebox[origin=l]{90}{~~~~\textbf{Supervised}} &  \subfloat{\includegraphics[width = 1.0in]{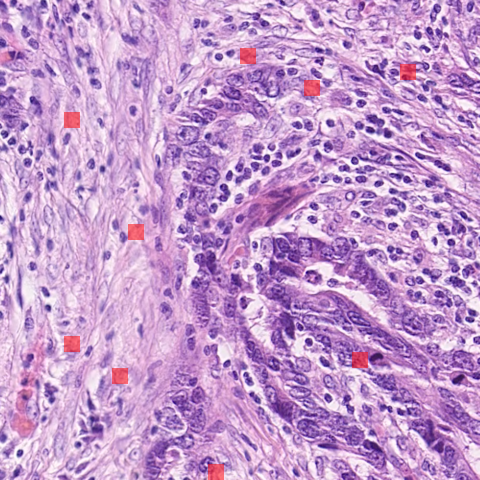}} &
    \subfloat{\includegraphics[width = 1.0in]{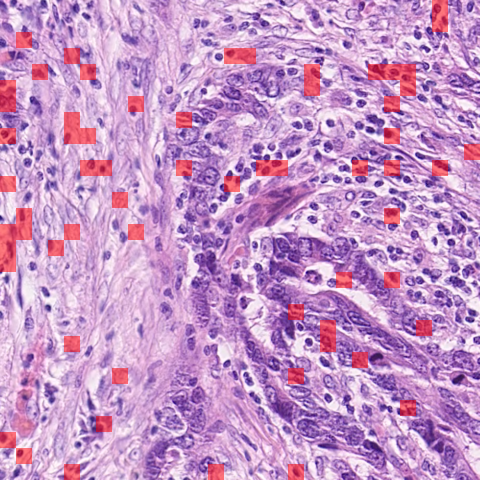}} &
    \subfloat{\includegraphics[width = 1.0in]{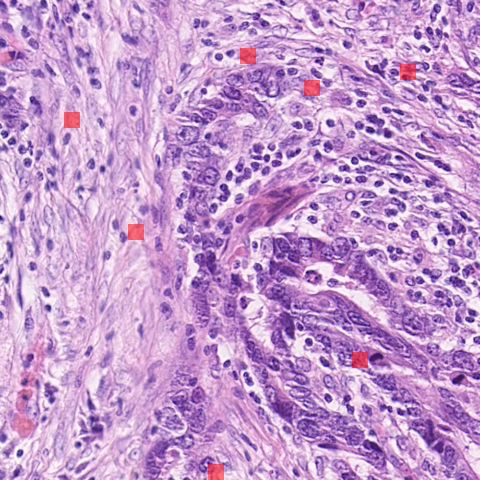}} &
    \subfloat{\includegraphics[width = 1.0in]{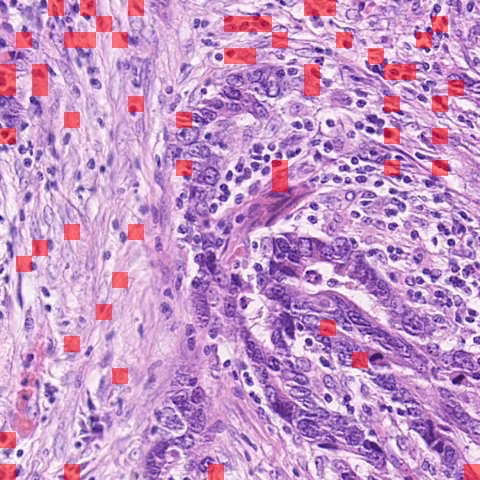}} &
    \subfloat{\includegraphics[width = 1.0in]{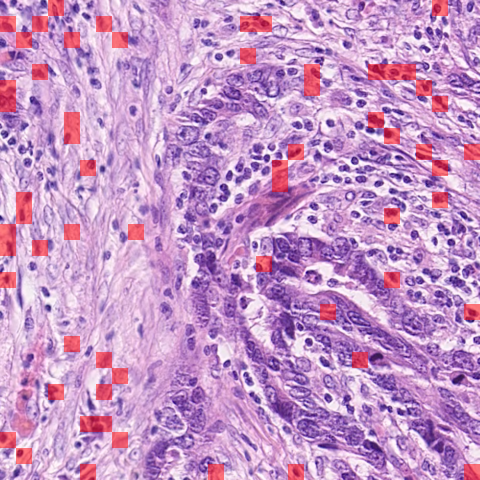}} &
    \subfloat{\includegraphics[width = 1.0in]{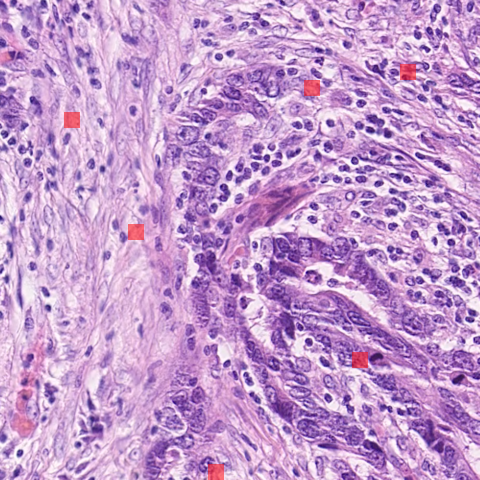}} \\ \\
\end{tabular}
\begin{tabular}{lcccccc}
\rotatebox[origin=l]{90}{\textbf{Self-Supervised}} & 
    \subfloat{\includegraphics[width = 1.0in]{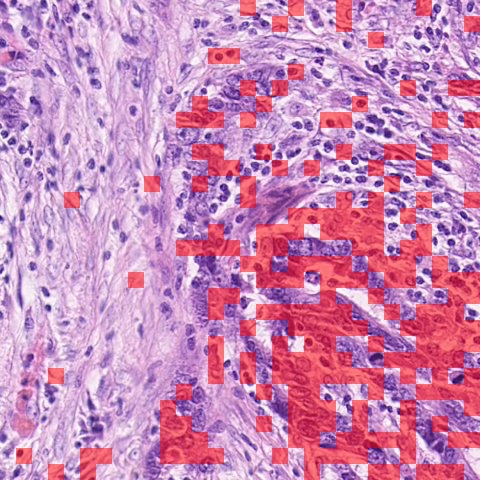}} &
    \subfloat{\includegraphics[width = 1.0in]{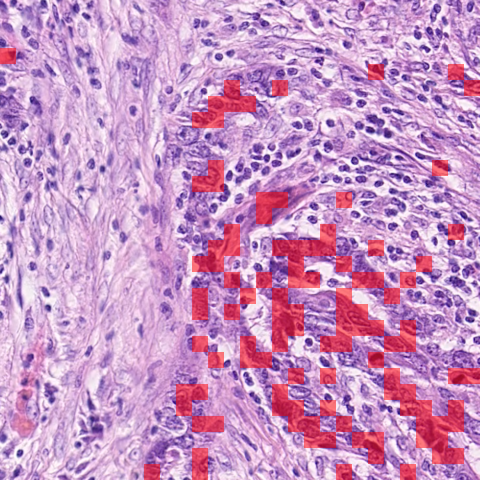}} &
    \subfloat{\includegraphics[width = 1.0in]{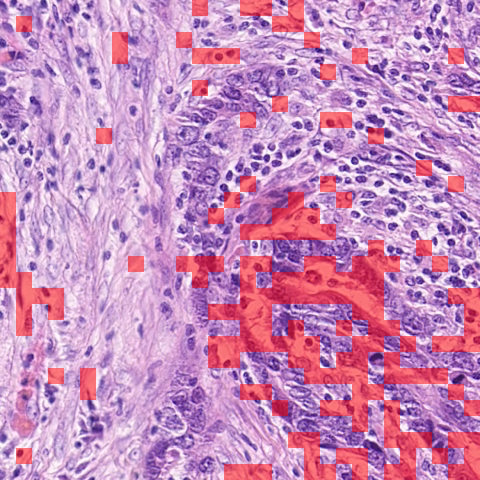}} &
    \subfloat{\includegraphics[width = 1.0in]{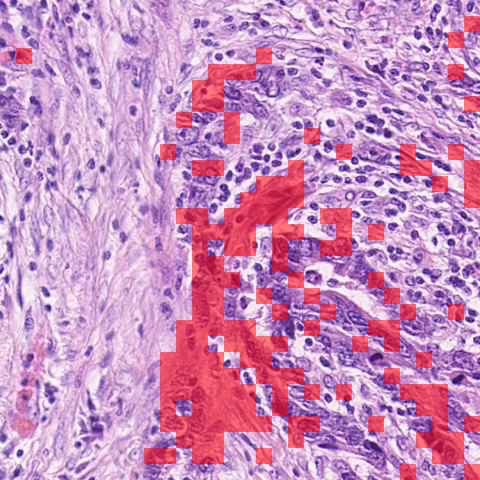}} &
    \subfloat{\includegraphics[width = 1.0in]{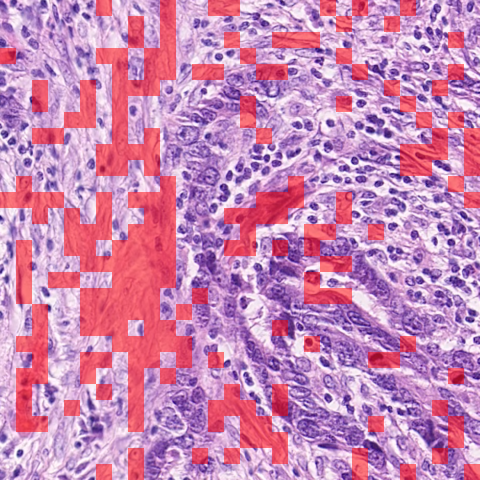}} &
    \subfloat{\includegraphics[width = 1.0in]{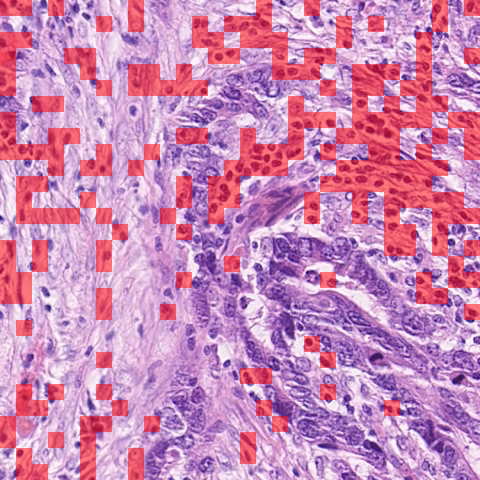}} \\ \hspace{-5.0mm} \vspace{-5.0mm}
\end{tabular}

\caption{\textbf{Visualizing multi-head self-attentions of ViT}. We visualize the attention map of several pre-trained ViT-S. Specifically, ViT-S has 6 attention heads. We visualize each head from the last layer of ViT. Our visualizations are presented in rows, with each row displaying attention maps alongside their corresponding overlayed image. The first two rows showcase the qualitative result of the supervised ViT pre-trained on ImageNet, while the next two rows display the qualitative result of the self-supervised ViT (DINO$_{p=16}$) pre-trained on TCGA. Note that, the input image is resized to 480 $\times$ 480 resolution, and overlaid in "red" are visual tokens whose attention weight > 0.5 and span 16 $\times$ 16 pixels.}
\label{supple:attention_maps}
\end{figure*}

As part of an effort to explore the potential of domain-aligned pre-training, we visualize the attention maps of self-supervised ViT and supervised ViT pre-trained on ImageNet.
Our results, as illustrated in \autoref{supple:attention_maps}, demonstrate that SSL ViT interestingly identifies and locates cells while also recognizing morphological phenotypes, which is aligned with recent observations\cite{chen2022scaling}. 
Specifically, attention heads 1 $\sim$ 4 attend to epithelial and inflammatory cells, whereas heads 5 $\sim$ 6 focus on fibroblast cells. 
In contrast, supervised ViT pre-trained on ImageNet fails to generate interpretable signals due to the domain gap, highlighting the effectiveness of domain-aligned pre-training in generating informative signals for downstream tasks.
We believe that this intriguing property can be leveraged to enable future potentials in the field of histopathology.

\section{Qualitative Results of Nuclei Instance Segmentation \color{blue}{(Section 5.2)}}

\begin{figure*}[t]
\center
\renewcommand{\arraystretch}{0.8}
\begin{tabular}{ccc}
 \textbf{Ground Truth} & \textbf{Supervised} & \textbf{\bt} \\
 \subfloat{\includegraphics[width = 2.0in]{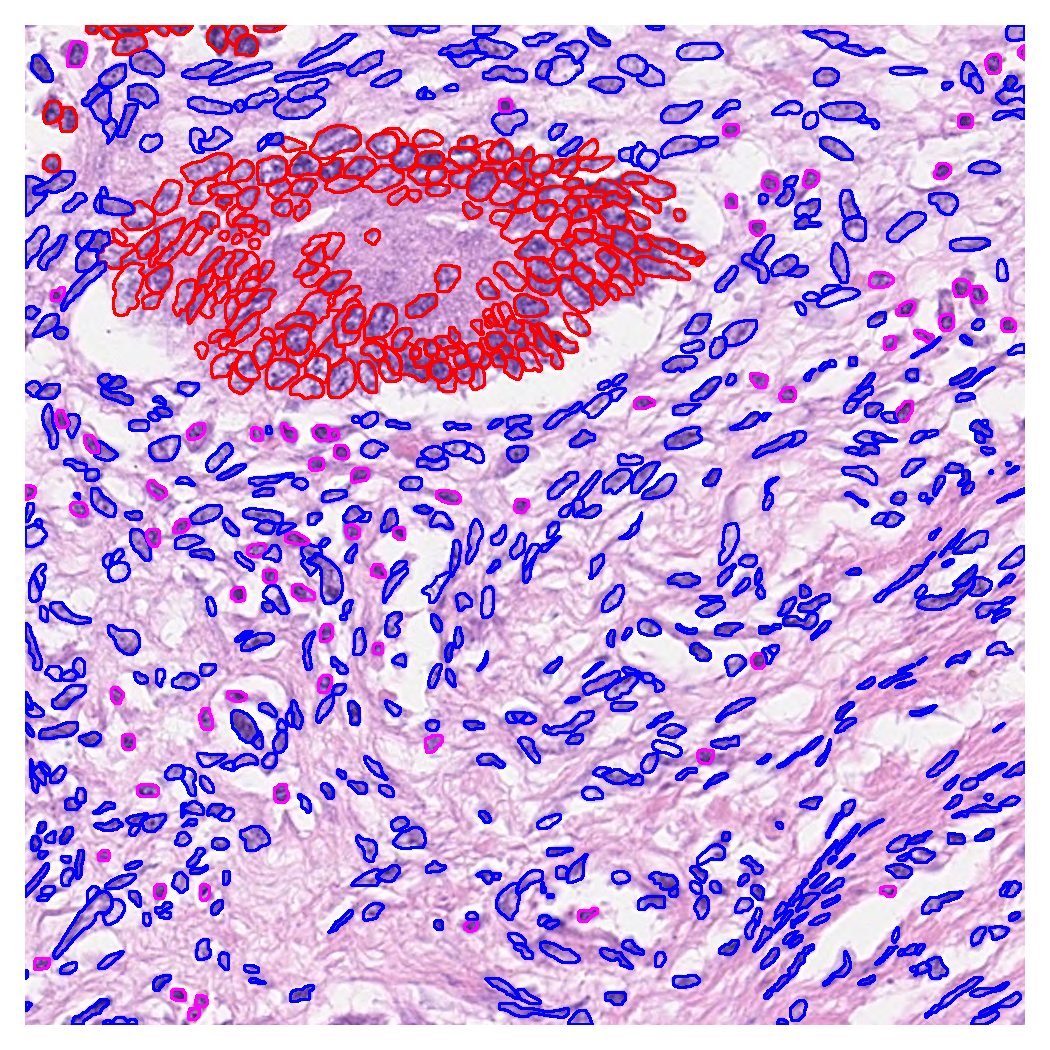}} &
\subfloat{\includegraphics[width = 2.0in]{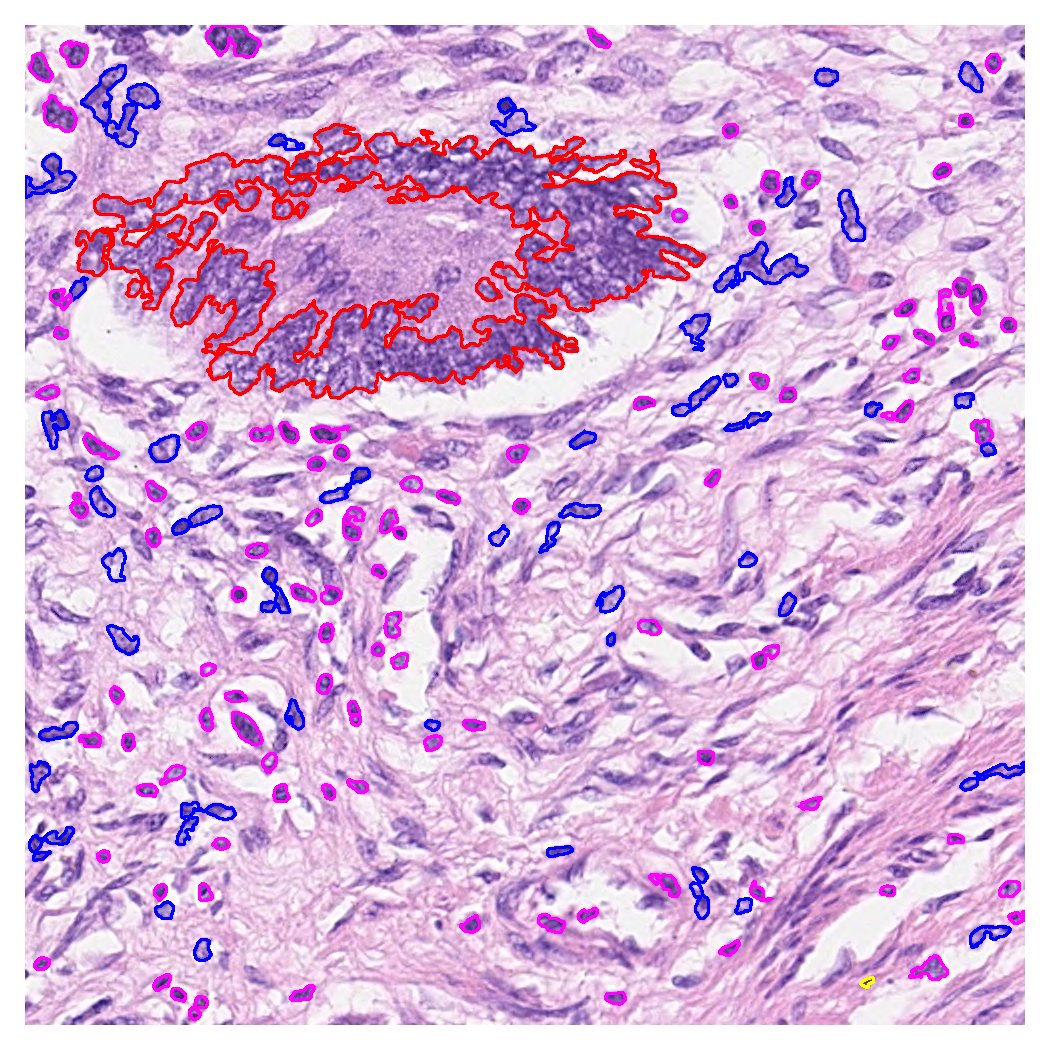}} &
\subfloat{\includegraphics[width = 2.0in]{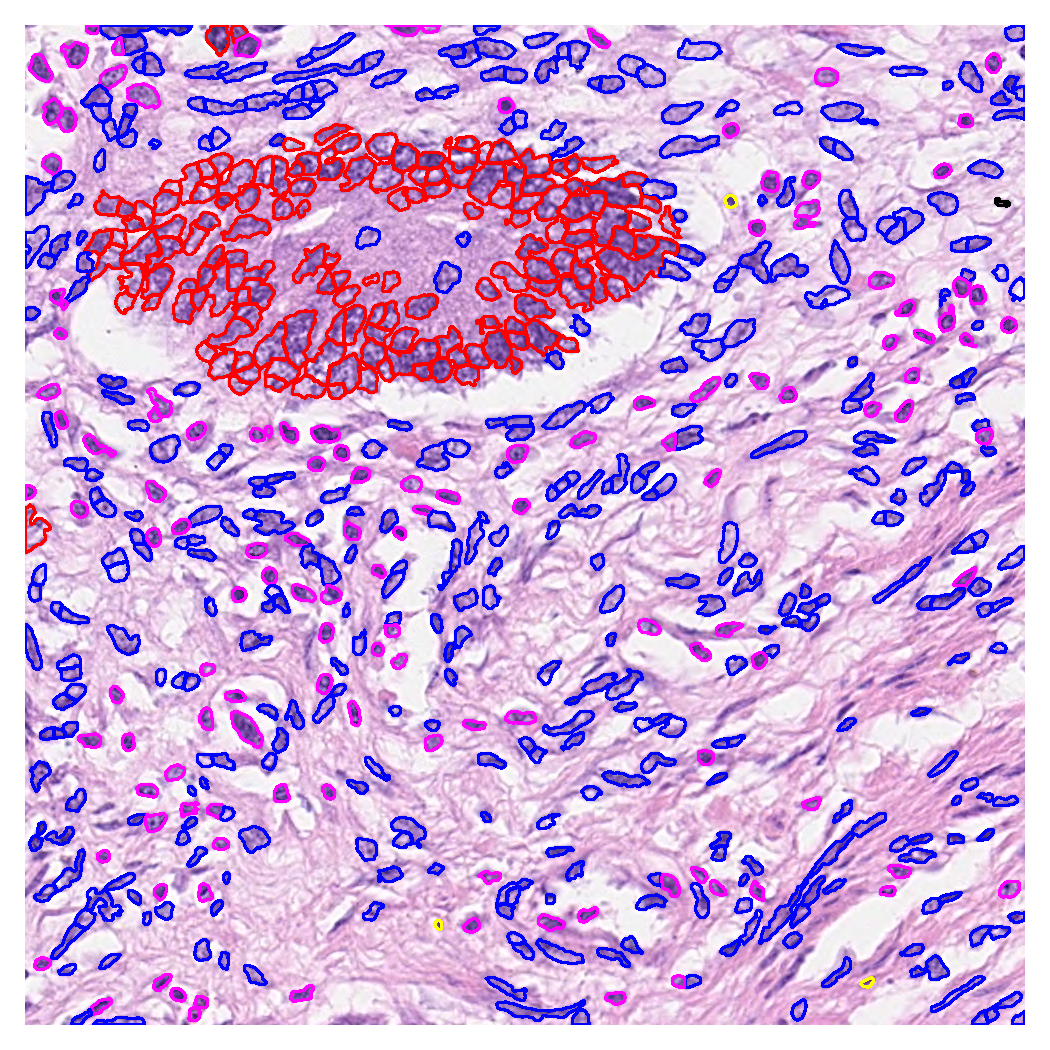}} \\ 
\subfloat{\includegraphics[width = 2.0in]{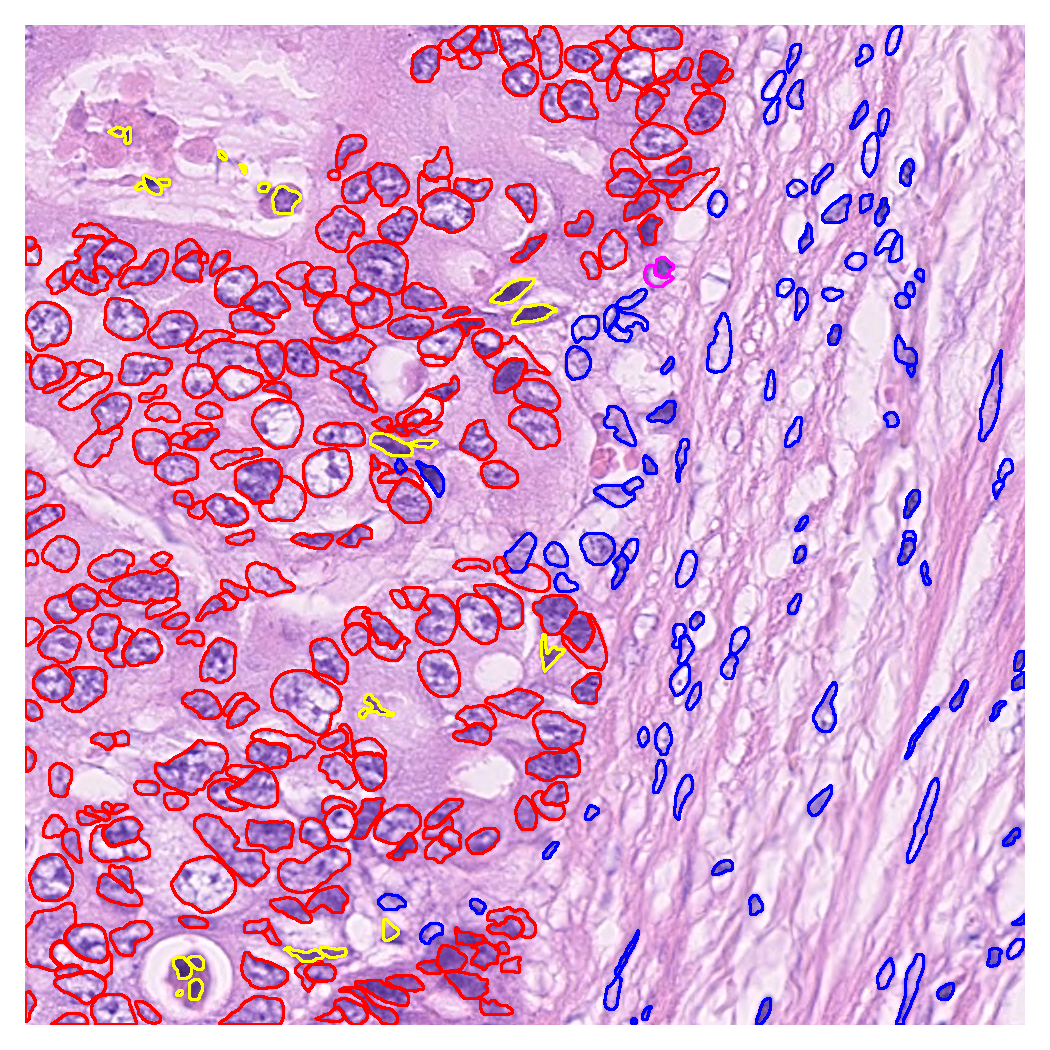}} &
\subfloat{\includegraphics[width = 2.0in]{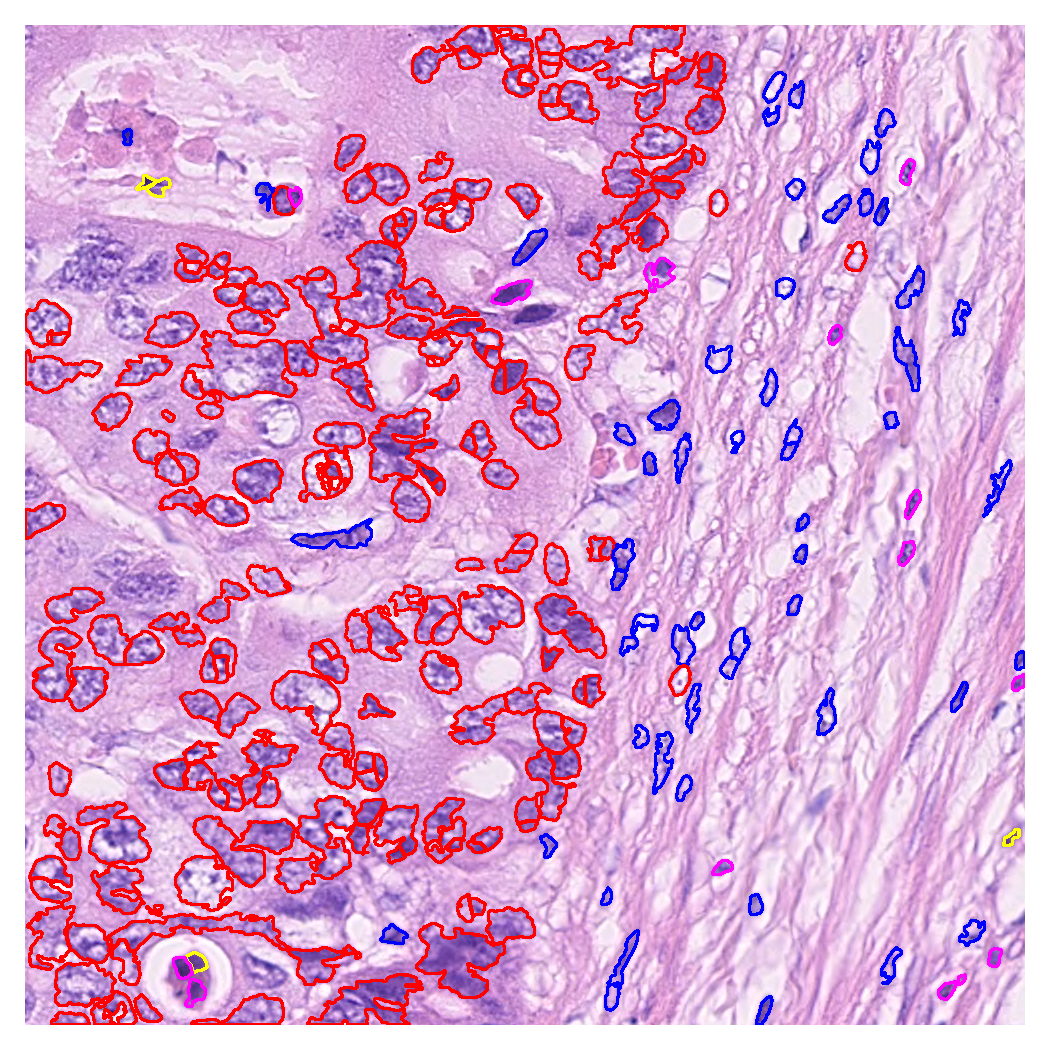}} &
\subfloat{\includegraphics[width = 2.0in]{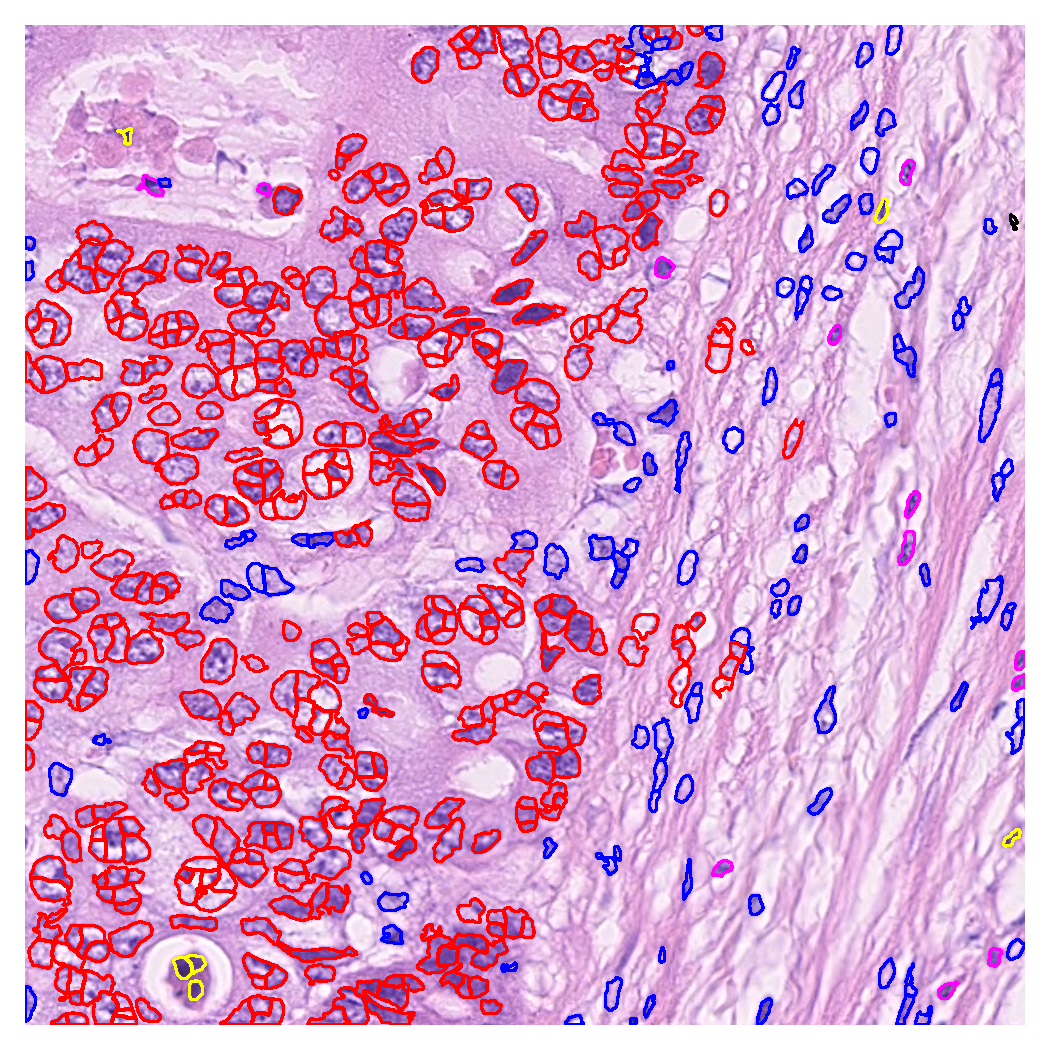}} \\
\end{tabular}
\caption{\textbf{Visualizing predictions of models.} We visualize the overlay predictions of different models on CoNSeP. A linear evaluation protocol is adopted to more directly assess the quality of the representations learned during pre-training. We selected the best-performing pre-trained model, \bt, based on the results of Table 4., obtained from the linear evaluation protocol.
We find that predictions from \bt are similar to the ground-truth, whereas the ``Supervised'' alternative produces poor nuclei boundaries and merges cells incorrectly.
}
\vspace{2mm}
\label{suppl:nuclei_seg_quality}
\end{figure*}

In order to perform a qualitative assessment of the effect of domain-aligned pre-training on nuclei instance segmentation, we compare the predictions of models using supervised ImageNet pre-training and self-supervised TCGA pre-training, adapted under the linear evaluation protocol.
The result presented in \autoref{suppl:nuclei_seg_quality} shows that domain-aligned pre-training can offer the benefit on downstream tasks effectively, resulting in capturing foreground cells and accurately classifying them, in contrast to the model trained using ImageNet pre-trained weight. 

\section{Slide-level Evaluation}

The slide-level classification task is outside of the scope of our work. Nonetheless, we conduct a preliminary experiment to demonstrate the usefulness of the features learned through SSL for this task, too. We train and test models for the classification of breast cancer metastases in WSIs, using the same configuration as CLAM~\cite{lu2021data} but on the Camelyon16~\cite{bejnordi2017diagnostic} dataset. To extract features from the WSIs, we use two pre-trained weights: ``Supervised (IN)'' and ``MoCo v2 (TC+TU)''. We find that models achieve an AUROC of 0.986 when using ``MoCo v2 (TC+TU)'' pre-trained weights, while models achieve an AUROC of 0.927 when using ``Supervised (IN)'' pre-trained weights. This result indicates that domain-aligned pre-training also can be beneficial to the slide-level task.

\end{document}